\documentclass[sigconf, table, dvipsnames]{acmart}

\usepackage{listings}
\usepackage{booktabs}
\usepackage{array}
\usepackage{multirow}
\usepackage{enumitem}
\usepackage{lineno}
\usepackage{subcaption}
\usepackage{tikz}
\usepackage{tabularx}
\usepackage{placeins} 
\usepackage{pdflscape} 

\usepackage{algorithm}
\usepackage{algpseudocode}

\usetikzlibrary{
  positioning,
  arrows.meta,
  shapes.geometric,
  fit,
  backgrounds,
  calc
}

\DeclareUnicodeCharacter{2264}{\ensuremath{\leq}}

\renewcommand\footnotetextcopyrightpermission[1]{} 

\AtBeginDocument{%
  \providecommand\BibTeX{{%
    \normalfont B\kern-0.5em{\scshape i\kern-0.25em b}\kern-0.8em\TeX}}}

\graphicspath{{./images/}} 
\begin{document}\emergencystretch=1.5em

\title{Evidential Deep Learning for Multi-Modal Anti-UAV Detection}

\author{Dmitry Golovchits}
\affiliation{%
  \institution{University of Amsterdam}
  \city{Amsterdam}
  \country{The Netherlands}
}
\email{dmitry.golovchits@student.uva.nl}

\author{Seyed Sahand Mohammadi Ziabari}
\authornote{Corresponding author.}
\affiliation{%
  \institution{University of Amsterdam}
  \city{Amsterdam}
  \country{The Netherlands}
}

\email{s.s.mohammadiziabari@uva.nl}

\author{Ali Mohammed Mansoor Alsahag}
\affiliation{%
  \institution{University of Amsterdam}
  \city{Amsterdam}
  \country{The Netherlands}
}
\email{a.m.m.alsahag@uva.nl}

\begin{abstract}

{
Anti-UAV systems increasingly fuse multiple sensors, yet their detection heads provide no per-modality reliability signal. This study evaluates whether evidential deep learning (EDL) heads, Dempster-Shafer (DS) evidence fusion, and uncertainty-driven temporal sensor gating improve anti-UAV detection, through a controlled ablation on three benchmarks: thermal tracking (AntiUAV600), RGB-audio-RF classification (TRIDENT), and RGB-IR tracking (MM-UAV). EDL heads improve classification accuracy over retrained sigmoid baselines in E2 (+4.8~pp, surviving a clip-clustered bootstrap, one-sided $p=0.011$) but not tracking accuracy in E1, where they increase the boxes output on target-absent frames. The remaining components fail their hypotheses. DS fusion does not outperform simple probability averaging. Dirichlet vacuity adds no ranking power beyond predictive entropy in the primary classification conditions; at the detection level it ranks errors about as well as the detection score over all detections (UAUC $\approx$0.96, a figure dominated by easy low-score errors) and nominally worse among confident detections (0.82 vs.\ 0.87). Temporal gating preserves accuracy only when nearly inactive and yields no realised latency saving, because every encoder runs before the gate decision. Evidential heads thus improve classification accuracy, while their uncertainty adds no usable signal beyond predictive entropy or the detection score; an exploratory crop-level control shows that isolating detected objects lowers vacuity's error ranking.
}

\end{abstract}

\keywords{Evidential Deep Learning, Dempster-Shafer Fusion, Anti-UAV Detection, Multi-modal Fusion, Uncertainty Quantification}


\fancyhead{}
\maketitle
\section{Introduction}
\label{sec:introduction}
Unmanned Aerial Vehicles (UAVs) are now ubiquitous in infrastructure inspection, logistics, and surveillance, yet their proliferation has created urgent security challenges. In 2025 alone, drone-related disruptions at European airports quadrupled year-on-year, forcing several temporary closures \cite{paternoster_fact-checking_2025}. The Russia-Ukraine war further demonstrated drones' military potential: Ukrainian manufacturers produced approximately 2.2 million UAVs in 2024 and were projected to exceed 4.5 million in 2025 \cite{nieczypor_game_2025, noauthor_russia-ukraine_2025}.

These threats share a common technical requirement: detecting, classifying, and tracking UAVs reliably across diverse and degraded sensor conditions. Current anti-UAV systems typically rely on single-modality detectors with standard confidence scores, offering no way to distinguish predictions backed by strong evidence from those arising from sensor noise or out-of-distribution inputs \cite{dong_securing_2025}. Multi-modal fusion improves robustness by combining complementary sensors, but most existing methods use learned gating or attention without an explicit per-sensor reliability model. A recent RGB--thermal extension of EDTC instead introduced evidential per-stream uncertainty with conflict-discounted belief fusion; however, its conflict-aware fusion was empirically indistinguishable from undiscounted averaging on the Anti-UAV benchmark because inter-modal conflict was nearly absent \cite{suttorp_uncertainty_2026}. This leaves open whether evidential fusion generalises across sensor combinations and task levels, and whether uncertainty can support practical sensor gating rather than only confidence reporting.

This study proposes EviDS-UAV (Evidential Dempster-Shafer fusion for anti-UAV) as a modular framework to investigate whether replacing conventional detection and fusion
components with evidential deep learning objectives and Dempster-Shafer (DS)
evidence fusion can improve detection accuracy, multimodal robustness, and
computational efficiency in anti-UAV systems. The evaluation focuses on
three complementary aspects: the effectiveness of Dempster--Shafer fusion
relative to standard multimodal fusion strategies, the role of evidential
uncertainty in reducing false positives and supporting detection--tracking
transitions, and the extent to which modality-specific uncertainty can inform
adaptive sensor utilisation.

Each proposed component is evaluated independently through controlled
ablation experiments against retrained baseline configurations rather than
being treated as an assumed design improvement. The results reveal an
asymmetric pattern. Evidential heads improve classification accuracy (E2) but
not tracking accuracy (E1), where they increase false boxes on target-absent
frames, whereas Dempster--Shafer fusion, vacuity-based filtering, and
temporal uncertainty gating do not outperform simpler alternatives.
Further analysis shows that predictive uncertainty ranks errors well in
balanced classification, and that at the detection level vacuity ranks
errors about as well as the detection score over all detections but no
better among confident ones. An exploratory crop-level control shows that
isolating detected objects lowers this ranking. These
findings therefore clarify both the potential and the limitations of
evidential uncertainty modelling for multimodal anti-UAV detection.

\section{Related Work}
\label{sec:related_work}

This section traces three research threads, anti-UAV detection and tracking, multi-modal fusion, and evidential uncertainty estimation, whose intersection defines the gap tested in this study.

The anti-UAV domain has evolved rapidly across detection, tracking, and classification, as surveyed by Dong et al. \cite{dong_securing_2025}. Early single-modality pipelines relied on RGB detectors such as YOLO variants \cite{ghazlane_real-time_2024, hao_high_2025} or infrared-specific architectures \cite{sun_multi-yolov8_2024}, achieving strong per-frame accuracy but with no mechanism to propagate reliability downstream \cite{dong_securing_2025}. A key advance, EDTC \cite{zhu_evidential_2023}, introduced evidential reasoning into the \emph{tracking} branch of a detection-tracking framework, switching to a global detector for re-acquisition when tracker uncertainty is high. However, its detector retains a standard sigmoid head, so detection-stage false positives propagate unchecked into the tracking loop. Subsequent extensions AMFT \cite{luo_adaptive_2025} and ADTC \cite{liu_learning_2025} improve switching through adaptive multi-feature fusion and augmented training respectively, yet both remain unimodal and do not introduce evidential learning into the detection branch. Related work outside anti-UAV has explored robustness in multi-object tracking through foundation-model feature distillation and multi-view aggregation under noisy, occluded observations \cite{faber_leveraging_2024, elchik_framework_2025}; these approaches strengthen representations or viewpoint integration rather than explicitly modelling predictive uncertainty.

\subsection{Multi-Modal Fusion for UAV Detection}

A parallel thread pursues multi-modal fusion to overcome single-sensor blind spots. Svanstr\"{o}m et al. \cite{svanstrom_dataset_2021} showed early that adding audio features to YOLO-based RGB or thermal detection reduces false alarms. The Anti-UAV benchmark \cite{jiang_anti-uav_2023} later formalised paired RGB-thermal evaluation, exposing persistent cross-modal misalignment and modality-specific degradation. Recent benchmarks broadened sensor coverage: MMAUD \cite{yuan_mmaud_2024} integrates stereo vision, LiDAR, radar, and audio arrays for 3D trajectory estimation; TRIDENT \cite{alla_trident_2025} synchronises RGB, audio, and RF streams for binary classification using both Late Fusion and Gated Multimodal Units (GMU); and MM-UAV \cite{xu_tri-modal_2025} pairs RGB, infrared, and event cameras for multi-object tracking via channel-attention fusion (ADFM). Methods built on these benchmarks consistently show that combining complementary sensors, audio-visual, RF-acoustic, and radar-camera, improves detection and classification robustness~\cite{xiao_av-dtec_2024, yang_av-fdti_2024, frid_drones_2024, dudczyk_multi-sensory_2022, kumar_uav_2024, wu_vehicle-mounted_2024, dong_securing_2025}. Most of these systems fuse modalities through learned gating, attention, or simple averaging without an explicit per-sensor reliability model. Suttorp et al. \cite{suttorp_uncertainty_2026} are a recent exception: they extend EDTC to RGB--thermal Anti-UAV perception using evidential uncertainty and conflict-discounted belief fusion, finding that multimodal fusion improves over either single stream but that conflict discounting itself is indistinguishable from undiscounted averaging when inter-modal conflict is nearly absent.

\subsection{Evidential Deep Learning and Uncertainty-Aware Fusion}

\looseness=-1 Independently, the uncertainty community has developed principled alternatives. Bayesian approximations via Monte Carlo dropout \cite{gal_dropout_2016} and deep ensembles \cite{lakshminarayanan_simple_2017} showed that networks can produce calibrated uncertainty estimates, while Guo et al. \cite{guo_calibration_2017} showed that modern networks are often poorly calibrated, fixable with post-hoc temperature scaling. Evidential deep learning (EDL) \cite{sensoy_evidential_2018} instead places a Dirichlet distribution over class probabilities, so a single forward pass yields both a prediction and an epistemic uncertainty estimate without ensemble overhead \cite{gao_comprehensive_2025}. DS theory has then been used to fuse per-modality evidence in a mathematically grounded manner. Li et al. \cite{li_stabilizing_2024} applied DS combination to stabilise multispectral pedestrian detection by integrating RGB and infrared evidence, while Shao et al. \cite{shao_dual-level_2024} introduced dual-level evidential fusion respecting both intra- and inter-modality uncertainty. In autonomous driving, Shao et al. \cite{shao_ua-fusion_2025} extended this with uncertainty-aware cross-modal attention for 3D object detection, and Durasov et al.~\cite{durasov_uncertainty_2024} found that standard EDL losses are not directly applicable to object detection, because extreme background-to-foreground imbalance biases the estimator toward underconfidence on positive detections. Cross-modality detection has also benefited: Sun et al. \cite{sun_drone-based_2022} and Zhao et al. \cite{zhao_uncertainty-aware_2024} use uncertainty-aware modules for visible-infrared fusion in drone-based vehicle detection, though neither uses a DS framework nor addresses detecting drones themselves.

\subsection{Research Gap}

A narrower but important gap remains. Recent work has now demonstrated uncertainty-aware evidential fusion for paired RGB--thermal anti-UAV perception \cite{suttorp_uncertainty_2026}, closing the earlier claim that this combination had not been studied in the domain. What remains unresolved is whether the observed behaviour generalises beyond one paired-vision benchmark and one conflict-discounting formulation: whether an evidential objective improves the detection head independently of fusion, whether evidence-based fusion outperforms simple averaging across heterogeneous RGB--audio--RF and RGB--IR settings, and whether modality-specific uncertainty can drive temporal sensor gating without sacrificing accuracy. EviDS-UAV is constructed as a controlled three-benchmark ablation to test these questions separately, as detailed in Section \ref{sec:methodology}.


\section{Methodology}
\label{sec:methodology}

EviDS-UAV is a modular method comprising three components, evidential detection heads, DS evidence fusion, and temporal sensor gating, that can be applied on top of existing anti-UAV baselines without modifying their feature extraction or tracking modules. All three baselines share a structural limitation: their detection or fusion weights are fixed or learned without any per-sensor reliability signal, so a degraded modality continues to influence the output with undiminished weight. EviDS-UAV is validated across three complementary experiments, each instantiating this limitation at a different level of the detection-fusion pipeline, to isolate the contribution of each component; Figure~\ref{fig:framework} gives an overview of the shared method and its instantiation across the three experiments.

\subsection{Shared Method}

\subsubsection{Evidential Detection Heads}
\label{sec:evidential_head}

\looseness=-1 In standard detection and classification pipelines, the final layer produces a point prediction via softmax or sigmoid activation, yielding a categorical probability vector $\mathbf{p} \in \mathbb{R}^K$ over $K$ classes. While often interpreted as confidence, these probabilities do not distinguish between cases where the model has strong evidence for a class and cases where it lacks evidence entirely — both can produce high-confidence outputs \cite{sensoy_evidential_2018, gal_dropout_2016}.

Following Sensoy et al. \cite{sensoy_evidential_2018}, the softmax classification layer is replaced with an evidential head that models a Dirichlet distribution over class probabilities. Given the feature representation $\mathbf{z}$ from a baseline's feature extractor, the evidential head computes:
\begin{equation}
    \mathbf{e} = f_\phi(\mathbf{z}), \qquad \boldsymbol{\alpha} = \mathbf{e} + 1,
    \label{eq:evidence}
\end{equation}
where $f_\phi$ is a linear layer followed by a softplus activation (the standard modern choice \cite{gao_comprehensive_2025}) ensuring $\mathbf{e} \geq 0$, and $\boldsymbol{\alpha} = (\alpha_1, \dots, \alpha_K)$ are the parameters of a Dirichlet distribution $\text{Dir}(\mathbf{p} \mid \boldsymbol{\alpha})$. The total evidence (Dirichlet strength) is $S = \sum_{k=1}^{K} \alpha_k$, from which the predicted class probability and epistemic uncertainty is derived:
\begin{equation}
\hat{p}_k = \frac{\alpha_k}{S}, \qquad u = \frac{K}{S}.
\label{eq:uncertainty}
\end{equation}
When the model observes familiar data, evidence $\mathbf{e}$ is large, $S \gg K$, and uncertainty $u \to 0$. For out-of-distribution or ambiguous inputs, evidence remains near zero, $S \approx K$, and $u \to 1$. This allows downstream components to distinguish between confident predictions and uninformed ones.

\subsubsection{Dempster-Shafer Evidence Fusion}
\label{sec:ds_fusion}

When multiple sensor modalities are available, each produces its own Dirichlet distribution via an independent evidential head. Rather than combining predictions through learned gating mechanisms (e.g. GMU \cite{alla_trident_2025}) or channel attention (e.g. ADFM \cite{xu_tri-modal_2025}), the DS combination rule is used to fuse them, which mathematically weights each modality's contribution by its evidence strength.

For each modality $m \in \{1, \dots, M\}$, the evidential head produces Dirichlet parameters $\boldsymbol{\alpha}^{(m)}$ with total evidence $S^{(m)}$. These are first converted to belief masses and an overall uncertainty mass following the framework of Li et al. \cite{li_stabilizing_2024}:
\begin{equation}
    b_k^{(m)} = \frac{e_k^{(m)}}{S^{(m)}}, \qquad u^{(m)} = \frac{K}{S^{(m)}},
    \label{eq:belief_mass}
\end{equation}
where $b_k^{(m)}$ is the belief assigned to class $k$ by modality $m$, and $u^{(m)}$ is the residual uncertainty. Note that $\sum_{k=1}^{K} b_k^{(m)} + u^{(m)} = 1$. For two modalities, the DS combination rule yields the fused belief:
\begin{equation}
    \hat{b}_k = \frac{1}{1 - C} \left( b_k^{(1)} b_k^{(2)} + b_k^{(1)} u^{(2)} + b_k^{(2)} u^{(1)} \right),
    \label{eq:ds_combination}
\end{equation}
where $C = \sum_{i \neq j} b_i^{(1)} b_j^{(2)}$ measures the conflict between the two modalities. The fused uncertainty is:
\begin{equation}
    \hat{u} = \frac{1}{1 - C} \, u^{(1)} \, u^{(2)}.
    \label{eq:ds_uncertainty}
\end{equation}
For $M > 2$ modalities (as in E2 with three sensors), the rule is applied iteratively: fuse modalities 1 and 2, then fuse the result with modality 3 \cite{li_stabilizing_2024, shao_dual-level_2024}. The DS rule has the property that a modality with high uncertainty ($u^{(m)} \to 1$, low evidence) contributes minimally to the fused output, while a modality with strong evidence dominates the decision. This is intended to contrast with learned fusion; whether it does so empirically is tested in E2 (Section \ref{sec:e2}).

\paragraph{Assumptions.} The DS combination rule assumes independent evidence sources and becomes unstable as inter-modal conflict $C \to 1$. Independence is approximately satisfied in E2 (RGB, audio, and RF measure distinct physical phenomena) but weaker in E3 (RGB and IR are correlated); fusion behaviour is therefore evaluated empirically. Under severe disagreement the fused class probabilities are driven toward $1/K$ \cite{li_stabilizing_2024}; the fused uncertainty mass (Equation~\ref{eq:ds_uncertainty}) does not increase under conflict, because it multiplies the two input uncertainties. The $(1-C)^{-1}$ normalisation is numerically unstable near $C=1$, so a fallback defers to the most certain single modality when $C > C_{\max}$. The distribution of $C$ is reported across all test samples; for E2, a canonical fusion ordering is fixed.

\subsubsection{Temporal Sensor Gating}
\label{sec:temporal_gating}

Inspired by UA-Fusion \cite{shao_ua-fusion_2025}, which mitigates sensor 
noise by dynamically re-weighting modality contributions through 
soft probabilistic attention, a complementary approach is adopted: 
a hard-gating mechanism that selectively disables entire modality 
streams to reduce computational cost. If modality $m$ exceeds an uncertainty threshold $\tau$ at time step $t$:
\begin{equation}
    g^{(m)}_{t+1:t+N} =
    \begin{cases}
        0 & \text{if } u^{(m)}_t > \tau_{\text{high}}, \\
        1 & \text{if } u^{(m)}_t < \tau_{\text{low}},\\
        g^{(m)}_{t} & \text{otherwise},
    \end{cases}
    \label{eq:gating}
\end{equation}
where $g^{(m)} = 0$ excludes modality $m$ from both feature extraction and fusion for the next $N$ time steps. In the E2 implementation evaluated here, a gated modality is excluded from fusion only; its encoder is still evaluated, so the efficiency results report projected rather than realised savings (Section~\ref{sec:evaluation}). Hysteresis is employed ($\tau_{\text{low}} < \tau_{\text{high}}$) to prevent oscillation when uncertainty hovers near a single boundary~\cite{zhu_evidential_2023, luo_adaptive_2025}. Note that the gating decision at time $t$ requires evaluating the modality at $t$; savings are realised only in frames $t+1$ through $t+N$. The constraint $\sum_{m} g^{(m)} \geq 1$ ensures at least one modality remains active, with an abstain fallback (the tracker coasting on its motion model) if post-fusion uncertainty remains high~\cite{xu_tri-modal_2025}. All thresholds and the suppression window $N$ are treated as hyper-parameters; their sensitivity is reported for E2 (Figure~\ref{fig:e2_gate_sweep}). The E3 implementation uses a streak-triggered variant of this rule: the gate closes only after $N$ consecutive frames exceed $\tau_{\text{high}}$ and remains closed until $u^{(m)} < \tau_{\text{low}}$, rather than for a fixed window. This adaptive-duration, noise-confirming form suits E3's per-frame, per-anchor density and variable-duration degradation events, whereas E2's coarser per-segment evaluation under steadier synthetic noise is matched by the fixed-window form. Whether this gate engages in E3 is evaluated empirically in Section~\ref{sec:results}.

\subsection{Training Objective}
\label{sec:loss}

Each evidential head is trained using the EDL loss from Sensoy et al. \cite{sensoy_evidential_2018}, which replaces the standard cross-entropy or binary cross-entropy loss used in the original baselines (\cite{zhu_evidential_2023}, \cite{alla_trident_2025}, \cite{xu_tri-modal_2025}). For a sample with one-hot ground truth label $\mathbf{y} \in \{0, 1\}^K$ and predicted Dirichlet parameters $\boldsymbol{\alpha}$, the EDL loss combines a Bayes risk term with a KL divergence regulariser:
\begin{equation}
    \mathcal{L}_{\text{EDL}} = \underbrace{\sum_{k=1}^{K} y_k \left( \psi(S) - \psi(\alpha_k) \right)}_{\text{Bayes risk (digamma variant)}} + \lambda_t \underbrace{\text{KL}\left[ \text{Dir}(\mathbf{p} \mid \tilde{\boldsymbol{\alpha}}) \;\|\; \text{Dir}(\mathbf{p} \mid \mathbf{1}) \right]}_{\text{KL regularisation}},
    \label{eq:edl_loss}
\end{equation}
where $\psi(\cdot)$ is the digamma function, $\tilde{\boldsymbol{\alpha}} = \mathbf{y} + (1 - \mathbf{y}) \odot \boldsymbol{\alpha}$ removes evidence for the correct class before computing the KL term, and $\lambda_t = \min(1, \, t / T_{\text{anneal}})$ is an annealing coefficient that gradually introduces the regulariser over $T_{\text{anneal}}$ training epochs (E1, E3) or optimiser steps (E2) to prevent premature evidence suppression during early training.

The Bayes risk term concentrates evidence on the correct class while the KL term drives evidence for incorrect classes toward the uniform prior, so the network learns to output high evidence for familiar, correctly classified inputs and low evidence for ambiguous or out-of-distribution samples. In the single-class detection settings (E1, E3), a background hypothesis is required: with $K{=}1$ the objective is identically zero, since $\psi(S) - \psi(\alpha_1) = 0$ and the KL term degenerates. In E1 the detector learns evidence for both background and UAV ($K{=}2$ evidential outputs per anchor), so vacuity is $u = 2/(\alpha_{\text{bg}} + \alpha_{\text{uav}})$. In E3 a fixed unit background evidence is appended ($\boldsymbol{\alpha} = [\alpha_{\text{uav}}, 1]$, giving $K{=}2$ effective classes), and detection-level vacuity is $u = 2/(\alpha_{\text{uav}} + 1)$, i.e.\ $K/S$ with the appended unit background evidence.

All other loss terms from the original baselines — including bounding box regression losses in E1 and E3, and any auxiliary losses — remain unchanged. In E2 and E3 only the classification component of each baseline's loss function is replaced by $\mathcal{L}_{\text{EDL}}$; in E1 the objectness and classification terms of YOLOv5 are jointly replaced by the two-class (background/UAV) $\mathcal{L}_{\text{EDL}}$.

\subsection{Baseline Selection and Justification}
\label{sec:baselines}
 
The three reproduced baselines and their selection rationale are summarised in Appendix Table~\ref{tab:baseline_selection}. All three were reproduced from their public codebases with corrections documented in Appendix~\ref{sec:apx:repro_dev}. A retrained variant of each baseline (matching the corrected training protocol) serves as the fair ablation comparator, ensuring that performance differences are attributable to the evidential formulation rather than to bug fixes or training data differences.

\subsection{Experiment 1: Evidential Detection in EDTC}
\label{sec:e1}
 
\subsubsection{Baseline Architecture}
 
EDTC~\cite{zhu_evidential_2023} is a detection-tracking collaboration framework for single-object UAV tracking in thermal infrared video. It consists of two independently trained stages:
 
\begin{enumerate}
    \item \textbf{Detector:} A YOLOv5s model with a standard sigmoid classification head, trained for 20 epochs on AntiUAV600 training sequences. The detector produces per-frame bounding box predictions with sigmoid confidence scores.
    \item \textbf{Tracker:} A CvT-based tracker with an evidential head that outputs epistemic uncertainty (vacuity). Stage~1 trains the CvT backbone for 500 epochs on five tracking datasets with ImageNet CvT-21 initialisation; Stage~2 fine-tunes the evidential head for 40 epochs on AntiUAV600.
\end{enumerate}
 
At inference, the system operates primarily in tracking mode. When the tracker's uncertainty exceeds a threshold ($\theta_{\text{eh}} = 0.2$, as reported optimal by Zhu et al.~\cite{zhu_evidential_2023}), or when the tracker classifies the region as background, the system switches to the YOLOv5s detector for global re-acquisition. Template update is disabled in the released implementation.
 
\subsubsection{EviDS-UAV Modification}

In EDTC the detector retains a standard sigmoid head, so detection-stage false positives propagate into the tracking loop unchecked; the tracker's evidential head can only react after a false-positive-initiated track has already begun. The YOLOv5s sigmoid objectness and classification outputs are replaced with a two-class evidential Dirichlet head (Section~\ref{sec:evidential_head}), trained with $\mathcal{L}_{\text{EDL}}$ (Section~\ref{sec:loss}). This creates a dual-evidential system where uncertainty is available at both detection and tracking stages. By setting an uncertainty threshold $\tau_{\text{det}}$, the evaluation tests whether detector vacuity can suppress false positives before they enter the tracker. The tracker and its existing evidential head remain unchanged.
 
\subsubsection{Data}
 
AntiUAV600 comprises 300 training sequences, 50 validation sequences (with ground truth), and 250 test sequences (ground truth withheld for server-side evaluation). All sequences are thermal infrared video with per-frame bounding box annotations and binary presence flags. The dataset is a multi-source superset assembled from at least three acquisition campaigns, identified by sequence naming conventions. Challenge attributes (Occlusion, Fast Motion, Scale Variation, Infrared Crossover, Dynamic Background Clusters, Target Scale) are annotated on a subset of sequences; per-attribute analysis is restricted to the 18 of the 50 validation sequences carrying non-empty attribute fields.
 
The majority of UAV bounding boxes fall below the COCO-small threshold ($<$1,024~px$^2$; median area $\approx$783~px$^2$), confirming a challenging small-object regime where the evidential head's uncertainty is hypothesised to correlate with target size.
 
\subsubsection{Training and Evaluation Protocol}
\label{sec:e1_protocol}
 
Table~\ref{tab:e1_protocol} specifies the training and inference pipelines for E1.
 
Evaluation uses the 50 official validation sequences as the held-out test set, since the competition server required for the 250-sequence test split is unavailable. This evaluation protocol differs from the published leaderboard and results are not directly comparable to test-set figures.

Accuracy (Acc) is computed with the released evaluation code for every condition: the mean per-frame score (IoU with the ground truth on target-present frames; 1 for an empty prediction and 0 otherwise on target-absent frames) minus $0.2\,m^{0.3}$, where $m$ is the fraction of frames on which the target is present but not found. Alongside standard metrics, Tracking on Absence (TrkOnAbs) is evaluated, defined as the fraction of absent frames where the tracker remains in control (uncertainty $< \theta_{eh}$) without delegating to the detector. A detection-level Uncertainty-Accuracy Area Under Curve (UAUC) is also evaluated to assess whether uncertainty ranks detection errors.
 
\subsubsection{Ablation Design}
 
Table~\ref{tab:e1_ablation} defines the four-condition ablation isolating the evidential head's contribution.
 
A change in TrkOnAbs is the primary success criterion for E1 because it directly measures the resilience of detection--tracking switching. A higher TrkOnAbs is a benefit only if the tracker outputs no box on target-absent frames, so the number of boxes output on absent frames is reported alongside it. Detection accuracy is therefore treated as a secondary outcome, while detector-level vacuity gating is evaluated as an empirical hypothesis through detector UAUC and attribute-stratified vacuity analyses. Temperature scaling, condition~(d), serves only as a calibration control. Because it is a monotonic transformation of the sigmoid outputs, it cannot alter the underlying switching decisions; consequently, its TrkOnAbs is identical to that of the baseline by construction. Any difference in TrkOnAbs between condition~(b) and condition~(a) can therefore be attributed to the evidential training objective rather than to post-hoc calibration.

\subsection{Experiment 2: Multi-Modal Evidential Fusion on TRIDENT}
\label{sec:e2}

\subsubsection{Baseline Architecture}

TRIDENT~\cite{alla_trident_2025} provides temporally synchronised RGB, audio, and RF data for binary UAV classification (drone vs.\ no-drone). Data is collected in 10-second clips, which are segmented into 0.25-second intervals for inference, each segment constitutes one classification decision; segments from the same clip are not independent (errors cluster in a few clips), so significance is assessed at the clip level. Each modality has its own feature extractor and sigmoid classification head, as detailed in Appendix Table~\ref{tab:e2_arch}.

The baseline tests two fusion strategies operating on frozen unimodal backbones: (1)~GMU with softmax gating weights over the modalities and a sigmoid output, and (2)~late fusion via a learned weighted combination of per-modality sigmoid outputs.

\subsubsection{EviDS-UAV Modification}

In TRIDENT both fusion strategies weight modalities with fixed or learned-but-uncalibrated coefficients, so under degradation a collapsed backbone (e.g.\ a video stream producing near-random outputs under noise) keeps its influence on the fused prediction. All three EviDS-UAV components are applied. First, each modality's sigmoid head is replaced with an evidential head (Section~\ref{sec:evidential_head}). Second, the GMU and late fusion modules are replaced with DS evidence fusion (Section~\ref{sec:ds_fusion}), combining the three modalities' Dirichlet evidence via the iterative combination rule. Third, temporal sensor gating (Section~\ref{sec:temporal_gating}) excludes a modality's evidence from fusion for subsequent segments when its uncertainty exceeds $\tau$ (encoder skipping is not implemented).

\subsubsection{Data}

TRIDENT comprises 277 independent 10-second clips (212 Train / 32 Validation / 33 Test) with binary labels: 159 Drone (57.4\%) and 118 Background. Each clip yields 40 evaluation segments of 0.25\,s, producing 1,320 test segments in total. The majority-class baseline accuracy at the segment level is 63.6\% (Drone class).

\looseness=-1 Noise degradation is applied programmatically at test time to audio and video streams; no separate degraded data files are stored. All E2 results in Section~\ref{sec:results} are reported under this test-time noise protocol (RF clean unless stated); a clean-test reference pass with degradation disabled is reported in Appendix Table~\ref{tab:e2_clean_test}. The public release contains only pre-computed RF spectrograms, raw I/Q data is unavailable, precluding faithful I/Q-level noise injection for RF.

\subsubsection{Training and Evaluation Protocol}
\label{sec:e2_protocol}

Table~\ref{tab:e2_protocol} specifies the training and inference pipelines.

\subsubsection{Ablation Design}

Table~\ref{tab:e2_ablation} defines the eight-condition ablation. To compare error-ranking quality under multi-modal uncertainty, an entropy-vs-vacuity UAUC is reported.

\subsection{Experiment 3: Multi-Modal Evidential Detection on MM-UAV}
\label{sec:e3}
 
\subsubsection{Baseline Architecture}
MMA-SORT~\cite{xu_tri-modal_2025} is a tri-modal (RGB, infrared, event) multi-object UAV tracking system. Dual-stream YOLOX detectors process RGB and IR independently. The Offset-Guided Adaptive Alignment module (OGAA, with deformable convolution and spatial transformer variants) spatially aligns the two streams, and the Adaptive Dynamic Fusion Module (ADFM) combines them via channel-attention weighting. Event camera data is used exclusively for motion-based identity association during tracking.

\subsubsection{EviDS-UAV Modification}
In MMA-SORT, ADFM's channel-attention weights carry no per-stream reliability signal, so a degraded stream, IR under thermal crossover, RGB in low light or smoke, keeps an attention weight uncorrelated with its reliability. Evidential heads replace the sigmoid classification outputs of both YOLOX streams. The OGAA spatial alignment module is retained unchanged in all conditions. The fusion point moves from ADFM's feature-level channel attention to decision-level DS evidence fusion operating on the per-stream Dirichlet outputs after the detection heads. Temporal gating between RGB and IR streams is optionally applied with the constraint that at least one remains active. The event-based tracking association is not modified.
 
\textbf{Architecture specificity.} ADFM operates at the feature level (Fusion0/1/2 in the FPN). When evidential heads are applied with ADFM retained (condition~c, Table~\ref{tab:e3_ablation}), ADFM fuses aligned features \emph{before} the per-modality evidential heads, so condition~(c) changes only the head formulation. When ADFM is replaced (conditions d--f), each stream's evidential head outputs $(\alpha, u)$ on the OGAA-aligned features and averaging or the DS rule fuses these at the decision level, replacing feature-level fusion with evidence-level fusion.

\subsubsection{Data}
MM-UAV provides 1{,}200 training and 121 test sequences with COCO-format RGB and IR annotations. No official validation split exists (the annotation files MM-UAV distributes as \texttt{val} cover the 121 test sequences), so a held-out 10\% of training sequences (a 1080/120/121 train/tune/test split, seed 42) is used for checkpoint selection and temperature fitting. Targets are extremely small (median bounding-box area $\approx$40~px$^2$, well below the COCO-small threshold), and per-frame annotation counts are highly correlated across RGB and IR, indicating that the two streams observe largely the same scene content, consistent with the weaker independence noted for the DS combination rule above and implying partial redundancy.

\subsubsection{Training and Evaluation Protocol}
\label{sec:e3_protocol}
Table~\ref{tab:e3_protocol} specifies the training and inference pipelines.

\subsubsection{Ablation Design}
Table~\ref{tab:e3_ablation} defines the seven-condition ablation isolating each component's contribution. Detection-level ECE is reported as a calibration diagnostic under every condition, and inter-modal disagreement is reported as $D=|p_{\text{rgb}}-p_{\text{ir}}|$ (Section~\ref{sec:evaluation}).
To examine whether detection-level uncertainty depends on the measurement context, an exploratory crop-level control is run on the saved E3 detections: image patches are extracted at detection locations ($1.5\times$ context margin, resized to $640\times640$ with aspect-preserving zero-padding) and evaluated through the same frozen evidential model with no retraining (the same crop is fed to both streams and vacuity is read from the RGB head). UAUC over matched versus unmatched crops is rank-based and prevalence-invariant, so no evaluation-set balancing is required. Crop-level vacuity UAUC is reported in aggregate and stratified by object scale, with sequence-clustered bootstrap confidence intervals where the stratum is large enough, for the averaging-trained and the DS-trained checkpoint (Appendix Table~\ref{tab:e3_crop_control}).

\subsection{Evaluation}
\label{sec:evaluation}
 
Across all experiments, evaluation targets three aspects aligned with the research sub-questions.
 
\subsubsection{Detection and Fusion Quality}
 
Each experiment reports task-specific performance metrics against its respective baseline: Acc for E1, accuracy and macro-F1 for E2, and MOTA/HOTA/IDF1 for E3. The ablation tables (Appendix \ref{sec:apx:ablations}) define the conditions under which each component's contribution is isolated.
 
\subsubsection{Uncertainty Quantification}

Expected Calibration Error (ECE) \cite{guo_calibration_2017} is reported under three conditions per experiment: (i)~the reproduced sigmoid baseline, (ii)~the baseline with post-hoc temperature scaling ($T$ fitted on the validation/tune split by NLL minimisation; E1: capped at 200), and (iii)~the evidential head without post-hoc correction. ECE serves only as a calibration diagnostic; the evidential claim is restricted to discrimination, i.e.\ the error-ranking quality measured by UAUC.
Reliability diagrams for the E1 conditions and the E2 baseline and evidential fusion conditions are provided in Appendix Figure~\ref{fig:apx:reliability}. For E2, ECE is computed both per-modality and on fused outputs.

UAUC evaluates whether uncertainty estimates rank detection errors. Sigmoid outputs yield only predictive entropy, whereas EDL additionally yields vacuity; although the two capture different dimensions of uncertainty, both are evaluated on a common AUROC-style ruler. As a fair-comparison control, the same score is used for the EDL conditions and the sigmoid baselines: predictive entropy in E2 and one minus the detection score in E1 and E3; vacuity-UAUC is additionally reported for the evidential models to isolate the ranking quality of epistemic mass.
 
\subsubsection{Efficiency}

For E2, operational efficiency is evaluated through two complementary analyses. First, hardware-level latency is measured with and without temporal gating under condition~(f) (Table~\ref{tab:e2_ablation}, Appendix~\ref{sec:apx:ablations}) on a single GPU. Because the benchmark evaluates every encoder before the gating decision and gated modalities are excluded only from fusion, realised encoder-time savings are not measured; savings are reported as projections from per-encoder times and observed skip rates. Second, an inference-only sweep over the gating hyperparameters $\tau_{\text{high}}$, $\tau_{\text{low}}$, and $N$ is performed on the frozen condition-(f) model to characterise the theoretical accuracy--skip-rate trade-off. This analysis tests whether any operating point achieves a meaningful per-modality skip rate without an unacceptable loss in accuracy and is reported as a trade-off curve.

For E3, the same efficiency question is examined at the stream level by testing whether per-stream uncertainty can suppress either the RGB or IR modality without degrading accuracy. Because the focus is on the feasibility of uncertainty-driven stream selection rather than realised hardware acceleration, the results report whether the gate engages and its effect on accuracy.

\subsubsection{Fusion Diagnostics}

\looseness=-1 For E2, the distribution of inter-modal conflict $C$ and the activation of the conflict-aware fallback ($C > C_{\max}$) are reported. For E3, with a single focal class and a fixed unit background evidence, $C$ is zero by construction, the conflict term requires competing class hypotheses to be simultaneously non-zero across both sources, so inter-modal disagreement is reported instead as $D=|p_{\text{rgb}}-p_{\text{ir}}|$ (Section~\ref{sec:results}). Accuracy under simultaneous all-sensor degradation is evaluated for E2.

\section{Results}
\label{sec:results}
\subsection{E1: Evidential Detection in EDTC}

\paragraph{Accuracy and switching.} Training the YOLOv5s detector with the EDL objective does not improve tracking accuracy over the sigmoid baseline under identical data and architecture: with the same evaluator for all conditions, EDL-270 reaches Acc 0.562 against 0.572 for Sigmoid-270 (Table~\ref{tab:e1_results}). EDL-270 roughly triples the tracker-on-absent rate (5.5\% to 17.4\%), but the tracker outputs a box on 282 of these 302 frames, so boxes on target-absent frames rise from 80 to 290. Since the tracker head is unchanged across conditions, this is an indirect, pipeline-level effect of the detector.
 
\paragraph{Calibration.} Det.\ ECE is dominated by detector-active frames without a detection (99\%), whose confidence is set to zero while the target is present on 87--89\% of them; it therefore reflects the detector's miss rate rather than the calibration of its scores, and no calibration difference between the sigmoid and evidential detectors is concluded. Temperature scaling (condition~(d)) lowers Det.\ ECE only because the fitted temperature reaches its search bound ($T=200$) and maps all confidences to $\approx 0.5$; as a monotonic rescaling of reported confidences it leaves switching, and therefore TrkOnAbs, identical to the untouched baseline.
 
\paragraph{Detector uncertainty.} \looseness=-1 Mean vacuity is similar on correct detections and false positives (0.44 vs.\ 0.45), and with only 5--6 false positives across all sequences neither vacuity nor confidence $1 - p_{\text{uav}}$ can be assessed as a false-positive signal (UAUC 0.54 vs.\ 0.76; Appendix Table~\ref{tab:e1_uncertainty_diag}). Attribute-stratified vacuity is computed over whole frames, which are predominantly background, and is uninformative about detection difficulty (Appendix Table~\ref{tab:e1_attr_detector}). The vacuity-thresholded condition~(c) lowers Acc relative to~(b) (0.547 vs.\ 0.562) but also reduces boxes on target-absent frames (117 vs.\ 290) and the tracker-on-absent rate (7.3\% vs.\ 17.4\%): $\tau_{\text{det}}$ suppresses 17 correct detections and one false positive.

\begin{table}
\centering
\caption{E1 results on 50 AntiUAV600 validation sequences (56,301 frames; 1,735 absent), focused view.
Acc is IoU-based accuracy with absence penalty (Section~\ref{sec:e1}).
Det.\ ECE and Trk.\ ECE are computed over detector-active and tracker-active frames respectively. \textbf{TrkOnAbs} = fraction of absent frames where the tracker remains in control (uncertainty $< \theta_{eh}$) without delegating to the detector; \textbf{AbsBox} = number of target-absent frames on which a box is output; TrkOnAbs is not bolded because a higher value is not better by itself. Det.\ ECE is dominated by detector-active frames without a detection (confidence 0) and mainly reflects the miss rate. Bold: best per column among conditions (a)--(d); the released checkpoint row is reference only. Overall (frame-weighted) ECE and throughput (constant at $\sim$20\,FPS) in Appendix Table~\ref{tab:e1_results_full}. All conditions share the same YOLOv5s + CvT-21 architecture.}
\label{tab:e1_results}
\resizebox{\columnwidth}{!}{%
\begin{tabular}{l c c c c c}
\toprule
\textbf{Condition} & \textbf{Acc} $\uparrow$ & \textbf{Det.\ ECE} $\downarrow$ & \textbf{Trk.\ ECE} $\downarrow$ & \textbf{TrkOnAbs} & \textbf{AbsBox} $\downarrow$ \\
\midrule
Released EDTC (\texttt{best.pt}) & 0.576 & 0.877 & 0.040 & 8.6\% & 132 \\
(a) Sigmoid-270 & \textbf{0.572} & 0.866 & 0.041 & 5.5\% & \textbf{80} \\
(b) EDL-270 & 0.562 & 0.875 & \textbf{0.036} & 17.4\% & 290 \\
(c) EDL-270 + $\tau_{\text{det}}\!=\!0.5$ & 0.547 & 0.888 & 0.040 & 7.3\% & 117 \\
(d) Sigmoid-270 + TempScale & \textbf{0.572} & \textbf{0.394} & 0.041 & 5.5\% & \textbf{80} \\
\bottomrule
\end{tabular}%
}
\end{table}
 
\subsection{E2: Multi-Modal Evidential Fusion on TRIDENT}

\begin{table}
\centering
\caption{E2 results on 1,320 TRIDENT test segments (33 clips $\times$ 40 segments), focused view. Mean $\pm$ sample standard deviation across noise seeds 42/123/456; ECE is top-label (15 bins). RF is evaluated clean-only throughout. UAUC measures how well uncertainty ranks classification errors, 0.5 is random, 1.0 is perfect. Predictive entropy ($\text{UAUC}_{\text{ent}}$) is computed identically for all models (from $\hat{p}$ for evidential heads); vacuity ($\text{UAUC}_{\text{vac}}$) is additionally reported for the evidential conditions only. Bold: best per column among fusion conditions (b)--(g). The five unimodal backbones (including the collapsed ResNet-10 Video) are in Appendix Table~\ref{tab:e2_results_full}.}
\label{tab:e2_results}
\resizebox{\columnwidth}{!}{%
\begin{tabular}{l l c c c c c}
\toprule
\textbf{Condition} & \textbf{Noise} & \textbf{Acc (\%)} $\uparrow$ & \textbf{Macro-F1} $\uparrow$ & \textbf{ECE} $\downarrow$ & \textbf{UAUC}$_{\text{vac}}$ $\uparrow$ & \textbf{UAUC}$_{\text{ent}}$ $\uparrow$ \\
\midrule
\multicolumn{7}{l}{\textit{Fusion baselines}} \\
\quad (b) Late Fusion & Noisy & 92.42 $\pm$ 0.20 & 0.920 $\pm$ 0.002 & 0.154 $\pm$ 0.001 & --- & 0.496 \\
\quad (c) GMU & Noisy & 61.44 $\pm$ 0.47 & 0.416 $\pm$ 0.009 & 0.317 $\pm$ 0.004 & --- & 0.506 \\
\midrule
\multicolumn{7}{l}{\textit{EviDS-UAV conditions}} \\
\quad (d) EDL + Average & Noisy & 97.20 $\pm$ 0.08 & 0.970 $\pm$ 0.001 & 0.254 $\pm$ 0.001 & 0.880 & \textbf{0.942} \\
\quad (e) EDL + DS & Noisy & \textbf{97.53} $\pm$ 0.17 & \textbf{0.973} $\pm$ 0.002 & 0.091 $\pm$ 0.003 & \textbf{0.935} & \textbf{0.942} \\
\quad (f) EDL + DS + Gating & Noisy & 91.82 $\pm$ 1.55 & 0.911 $\pm$ 0.018 & \textbf{0.054} $\pm$ 0.007 & 0.869 & 0.813 \\
\midrule
\multicolumn{7}{l}{\textit{Full-degradation foil}} \\
\quad (d$'$) EDL + Average & Noisy (all) & 65.73 $\pm$ 0.27 & 0.449 $\pm$ 0.009 & 0.148 $\pm$ 0.004 & 0.726 & 0.727 \\
\quad (g) EDL + DS$^\ddagger$ & Noisy (all) & 63.91 $\pm$ 0.04 & 0.397 $\pm$ 0.001 & 0.328 $\pm$ 0.002 & 0.812 & 0.778 \\
\bottomrule
\multicolumn{7}{l}{\footnotesize $^\ddagger$(g) collapses to majority-class prediction (Drone on 99.7--99.8\% of segments); its UAUC is a majority-class artefact.} \\
\end{tabular}%
}
\end{table}

\paragraph{Accuracy.} The evidential heads, trained on the frozen retrained backbones, deliver a clear accuracy gain: EDL + Average reaches 97.20\%, +4.78~pp over the strongest correctly-reproduced baseline (Late Fusion, 92.42\%), at macro-F1 0.970 (Table~\ref{tab:e2_results}). Because the 1{,}320 segments derive from only 33 independent clips, significance is assessed with a paired clip-level bootstrap ($n{=}33$, seed-42 outputs): the EDL-plus-average gain over Late Fusion survives clustering (clip-level point estimate 4.99~pp versus the 4.78~pp three-seed segment-level difference; 95\% CI $[+0.61, +11.06]$~pp, one-sided $p=0.011$), and the EDL-plus-DS gain survives likewise (pooled over the three seeds: clip-level $\Delta = +5.09$~pp versus the $5.11$~pp segment-level difference; 95\% CI $[+0.15, +11.52]$, one-sided $p=0.021$, uncorrected). The GMU baseline collapses to 61.44\% --- below the 63.6\% majority-class floor --- predicting Drone on 94.2\% of segments, and is treated as cautionary only (Appendix~\ref{sec:apx:repro_dev}). Late Fusion is the reference point for all fusion comparisons.

\paragraph{Fusion.} The DS rule does not justify its complexity. Under noisy-but-informative conditions, DS (e, 97.53\%) edges ahead of averaging (d, 97.20\%) by $+0.33$~pp, a difference far from significance under the clip-clustered bootstrap (95\% CI $[-0.68, +1.67]$~pp, one-sided $p=0.332$, pooled over the three seeds), and is better calibrated (top-label ECE 0.091 vs.\ 0.254, Table~\ref{tab:e2_results}). Under full degradation both rules collapse toward the majority class (63.6\%): (d$'$) predicts Drone on 97--98\% of segments and (g) on 99.7--99.8\% (65.73\% vs.\ 63.91\% accuracy). Inter-modal conflict $C$ peaks at 0.893 under the standard noisy protocol, below the 0.9 fallback threshold; under the all-noise foil it reaches 0.907 and the fallback fires on a handful of segments, yet (g) still collapses: the formalism executes correctly, its conflict machinery activates only when fusion is already lost, and it confers no advantage over the parameter-free average. The DS-versus-average ordering is itself sensitive to the belief-mass construction, but lies within the bootstrap interval under both constructions tested, so the absence of a detectable difference between the two rules is robust to this choice (Appendix~\ref{sec:apx:unc_diag}).

\paragraph{Uncertainty ranking and calibration.} The EDL conditions reach a higher entropy UAUC than the sigmoid baseline (0.942 for both DS and Average, versus 0.496 for Late Fusion), but the errors concentrate in 4--8 of the 33 clips and the clip-level difference is not significant (paired 95\% CI $[-0.05, +0.69]$, seed 42); in the primary fusion conditions predictive entropy from the Dirichlet mean matches or exceeds vacuity (Table~\ref{tab:e2_results}), so vacuity adds no usable ranking power beyond $\hat{p}$. In the degraded conditions the ordering nominally reverses (0.869 vs.\ 0.813 under gating; 0.812 vs.\ 0.778 under all-noise), but (g)'s ranking is a majority-class artefact and both gated values trail the ungated conditions. Because the EDL model is also more accurate, the UAUC gap reflects the joint effect of objective and accuracy, not vacuity specifically. On calibration, temperature-scaled Late Fusion (top-label ECE 0.057) is as well calibrated as the best evidential conditions (gated 0.054, DS 0.091), while evidential averaging is the least calibrated fusion rule (0.254; Table~\ref{tab:e2_tempscale}); as a monotonic transform, temperature scaling leaves the baseline's error ranking unchanged by construction (UAUC 0.496).

\paragraph{Gating.} Temporal gating yields no operationally useful trade-off, and the configuration reported as condition~(f) sits in the degradation regime. Its hysteresis thresholds ($\tau_{\text{high}}=0.35$, $\tau_{\text{low}}=0.05$, $N=10$, tuned on the validation split) drive a 27--32\% mean skip rate across seeds (16--58\% per modality; Table~\ref{tab:e2_efficiency} gives seed 42) and a $\sim$5.7~pp accuracy loss against ungated DS.
The full sweep over 664 configurations (Figure~\ref{fig:e2_gate_sweep}) shows the trade-off is sharply non-monotonic: light gating ($\tau_{\text{high}}=0.4$, $\tau_{\text{low}}=0.3$, $N=3$; 3.2\% skip) \emph{gains} $+0.9$~pp over ungated DS (98.3\% vs.\ 97.42\%, seed 42; the best of 664 configurations evaluated on the test split, hence optimistic); configurations with 8--13\% skip range from $+0.7$ to $-3.9$~pp (median $+0.7$), and every configuration beyond $\sim$13\% skip loses at least 3~pp. Even on clean inputs the gate keeps the audio stream off for 53.9\% of segments, costing accuracy (97.35\% vs.\ 99.02\%; Appendix Table~\ref{tab:e2_clean_test}). Projected latency savings scale with skip rate but are not realised, because the benchmark evaluates every encoder before the gate decision (Table~\ref{tab:e2_efficiency}).

\subsection{E3: Multi-Modal Evidential Detection on MM-UAV}

\begin{table}
\centering
\caption{E3 tracking and detection-level uncertainty on MM-UAV (121 test sequences), focused view. MOTA is reported per stream; the decision-level fusion conditions (d)--(f) return a single fused detection in RGB coordinates, so per-stream IR metrics are undefined. ECE and UAUC are detection-level (RGB stream, post-NMS, IoU $\geq$ 0.5 to GT, all detections above the 0.001 analysis floor); UAUC$_{\text{vac}}$ and UAUC$_{1-\text{conf}}$ rank errors by vacuity and by one minus the detection score. Single training seed per condition. Bold: best per column among operating conditions (a), (b), (d)--(g). Full HOTA/IDF1/IDs and per-stream metrics, plus the conf$=$0.1 diagnostic row (c$'$), in Appendix Table~\ref{tab:e3_results_full}.}
\label{tab:e3_results}
\resizebox{\columnwidth}{!}{%
\begin{tabular}{l c c c c c}
\toprule
\textbf{Condition} & \textbf{RGB MOTA} $\uparrow$ & \textbf{IR MOTA} $\uparrow$ & \textbf{ECE} $\downarrow$ & \textbf{UAUC}$_{\text{vac}}$ $\uparrow$ & \textbf{UAUC}$_{1-\text{conf}}$ $\uparrow$ \\
\midrule
(a) MMA-SORT DefConv         & \textbf{64.28} & \textbf{80.74} & 0.051 & ---   & 0.959 \\
(b) MMA-SORT STN             & 64.22          & 79.09 & 0.052 & ---   & \textbf{0.964} \\
(c) Evidential + ADFM        & 0.02           & 1.22  & 0.475 & 0.817 & 0.893 \\
(d) Evidential + Average     & 55.91          & ---   & \textbf{0.035} & 0.960 & 0.957 \\
(e) Evidential + DS Fusion   & 56.57          & ---   & 0.036 & \textbf{0.972} & 0.959 \\
(f) Evidential + DS + Gating & 56.57          & ---   & 0.036 & \textbf{0.972} & 0.959 \\
(g) Baseline + TempScale     & 62.16          & 79.88 & 0.058 & ---   & 0.944 \\
\bottomrule
\end{tabular}%
}
\end{table}

\paragraph{Baselines.} The retrained baselines establish two reference points and a cross-modal asymmetry. MMA-SORT with deformable-convol\-ution alignment~(a) reaches RGB MOTA 64.28 and IR MOTA 80.74; the spatial-transformer variant~(b) is close to it on every metric (Table~\ref{tab:e3_results}; full HOTA/IDF1/IDs breakdown in Appendix Table~\ref{tab:e3_results_full}). Infrared tracking exceeds RGB by roughly 15--16 MOTA points across both, confirming the thermal stream as the more discriminative modality for small-UAV detection in this benchmark.
 
\paragraph{Confidence-scale collapse.} Replacing the sigmoid heads with evidential heads while retaining ADFM (condition~c) collapses tracking at the published detection threshold, with RGB MOTA falling to 0.02 and IR to 1.22. The detections are not lost: at a negligible threshold ($\text{conf}=0.001$) the evidential head produces 23{,}743 detections that overlap a ground-truth box (Table~\ref{tab:e3_confsweep}), but almost none survive tracking: recall is 0.3\% at $\text{conf}=0.1$ (condition~c$'$, where new tracks still require a score above 0.7) and 0.1\% at $\text{conf}=0.3$ (condition~c), while the baseline retains 78.5\% recall at $\text{conf}=0.3$. A plausible explanation is that the threshold $\text{obj}\times\text{cls}\geq0.3$ is calibrated to sigmoid magnitudes, whereas the Dirichlet-mean scores $\hat{p}_k=\alpha_k/S$ are compressed toward $1/K$; however, conditions~(d) and~(e) use the same scores and do not show this collapse, so the cause was not isolated.
 
\paragraph{Decision-level fusion.} Moving the fusion to decision level (conditions~d, e) restores tracking to within range of the baseline without matching it. Probability averaging~(d) reaches RGB MOTA 55.91 and DS combination~(e) 56.57, both retaining RGB box geometry and discarding the IR coordinate frame, so per-stream IR metrics are undefined for these conditions. DS sits 0.65 MOTA above averaging (single training seed, no interval) with comparable HOTA, IDF1, and calibration. Both fall below the RGB baseline (64.28) and well below IR (80.74): averaging improves recall over the RGB baseline (80.8\% versus 78.5\%, 653 fewer misses) but introduces 2{,}851 additional false positives and 147 additional identity switches, and the false-positive cost dominates (Table~\ref{tab:e3_confsweep}). A possible cause is that aligned RGB and IR boxes do not coincide exactly, creating duplicate and offset tracks; this was not tested. With a single focal class the DS conflict $C$ is structurally zero, and the disagreement proxy $D=|p_{\text{rgb}}-p_{\text{ir}}|$ correlates only weakly with the fused uncertainty ($r=-0.14$), so the conflict-driven down-weighting that distinguishes DS from averaging never activates.
 
\paragraph{Detection-level uncertainty.} Over all detections, vacuity ranks errors well in the decision-level conditions (UAUC 0.960 for~d, 0.972 for~e), on par with one minus the detection score (0.957, 0.959) and with the scores of the sigmoid baselines (a) and (b) (0.959--0.964, Table~\ref{tab:e3_results}); this aggregate is dominated by easy low-score errors (85\% of detections are errors). Among detections above the operating threshold (score $\geq 0.3$; subset specified before retraining, after an exploratory look at the v1 checkpoints), vacuity still ranks errors above chance but nominally below the detection score: 0.819 [0.766, 0.858] versus 0.868 [0.830, 0.897] for~(d) and 0.818 [0.769, 0.855] versus 0.868 [0.832, 0.896] for~(e) (sequence-clustered 95\% CIs). With ADFM~(c), vacuity trails the score overall (0.817 vs.\ 0.893) and is at chance among confident detections (0.492). Vacuity therefore offers no ranking advantage over the detection score for detector-level gating in E3.
 
\paragraph{Calibration.} As a calibration diagnostic, the decision-level evidential conditions~(d, e) reach detection-level ECE 0.035--0.036, below the retrained baselines (0.051--0.052) and the temperature-scaled baseline (0.058), whereas the evidential-head-with-ADFM condition~(c) is poorly calibrated (ECE 0.475, Table~\ref{tab:e3_results}).
 
\paragraph{Gating.} Temporal gating (condition~f) is non-operational: its tracker output is identical to that of~(e). The gate compares, per frame and stream, the mean vacuity over anchors with objectness above 0.1 against hysteresis thresholds ($\tau_{\text{high}}=0.65$, $\tau_{\text{low}}=0.40$, $N=8$); the gate-off condition is never met, no stream is skipped, and~(f) reduces exactly to~(e). Anchor-level vacuity is strongly bimodal (for the RGB stream of~(e), 42\% of anchors below 0.02 and 37\% above 0.98), leaving little middle range for threshold-based gating.
 
\paragraph{Temperature scaling.} Temperature scaling of the baseline (condition~g) confirms the detector is already well-calibrated: it slightly worsens ECE (0.058 vs.\ 0.051), with a $T=0.626$ fitted on the tune split that sharpens rather than softens the scores. Because the 0.3 tracking threshold and the 0.001 analysis floor are applied after scaling, it also changes tracking (RGB MOTA 62.16 versus 64.28) and the set of analysed detections (41{,}000 versus 102{,}801).

\paragraph{Crop-level control.} In the exploratory crop-level control (Section~\ref{sec:e3}), vacuity ranks errors above chance on isolated detection crops but worse than over all detections of the same model: UAUC 0.714 (sequence-clustered 95\% CI $[0.684, 0.736]$) for the averaging-trained checkpoint and 0.605 ($[0.554, 0.646]$) for the DS-trained checkpoint, against 0.960 and 0.972 over all detections. The small-object stratum that dominates the benchmark follows the aggregate (0.732 and 0.583); the medium and large strata are sparse (Appendix Table~\ref{tab:e3_crop_control}). Isolating the detected object therefore lowers vacuity's error ranking.


\section{Discussion}
\label{sec:discussion}

The results reveal a clear asymmetry across the three investigated components. Dempster--Shafer (DS) fusion provides no advantage over standard fusion. In E2, its performance is statistically indistinguishable from probability averaging, and in E3 the two rules differ by less than one MOTA point in a single training seed (Section~\ref{sec:results}). These findings indicate that the additional uncertainty-aware fusion mechanism does not translate into improved predictive performance under the evaluated conditions.

The evidential training objective shows a mixed effect. In E1, where the conditions differ only in the detector's objectness and classification terms, evidential deep learning (EDL) does not improve tracking accuracy (Acc 0.562 versus 0.572) and roughly triples the tracker-on-absent rate, which here means more boxes on target-absent frames (290 versus 80; Section~\ref{sec:results}). In E2, evidential heads trained on the frozen backbones improve classification accuracy by 4.8~pp over Late Fusion. The evidential uncertainty measure itself adds little: in E2 vacuity ranks errors no better than the predictive entropy of the Dirichlet mean, in E3 it ranks detection errors about as well as the detection score overall and nominally worse among confident detections, and E1 has too few false positives to assess it. The results therefore separate two effects that are often treated jointly in evidential formulations: the effect of training evidential heads and the inference-time information provided by vacuity.

Uncertainty-driven sensor management likewise provides no meaningful efficiency benefit under the evaluated settings. In E2, gating does not reduce latency because the benchmark evaluates every modality encoder before the gating decision is made. Moreover, accuracy deteriorates once the skip rate moves beyond a narrow low-skip regime. In E3 the gate never engaged; anchor-level vacuity is strongly bimodal, leaving little middle range for threshold-based sensor selection. Taken together, evidential heads improved classification accuracy in E2 but not tracking accuracy in E1, and neither DS-based uncertainty fusion nor vacuity-driven sensor gating provides a consistent advantage in the examined multimodal detection settings.

\subsection{Evidential Uncertainty at the Detection versus Classification Level}
 
The central finding of this study is that evidential uncertainty adds no error-ranking information beyond simpler signals: predictive entropy in classification and the detection score in detection. In E2's balanced per-segment binary classification, predictive entropy from the Dirichlet mean ranks errors well (UAUC 0.94), although vacuity adds nothing beyond it and the errors concentrate in a few clips, so the ranking advantage over the sigmoid baseline is not established at the clip level. At the E3 detection level, vacuity over all detections ranks errors about as well as the detection score (UAUC 0.96--0.97 versus 0.96), but this aggregate is dominated by easy low-score errors. Among detections above the operating threshold, vacuity still ranks errors above chance but nominally below the detection score (0.82 versus 0.87, overlapping intervals), and with ADFM it is at chance. E1 has too few false positives (5--6) to assess either signal.
 
Standard EDL is known to be unreliable under the extreme class imbalance of object detection~\cite{durasov_uncertainty_2024, gao_comprehensive_2025}. The results here are consistent with a milder form of this: neither vacuity nor the detection scores fail in aggregate (UAUC 0.96--0.97 for the decision-level conditions and the sigmoid baselines), but among confident detections the signals of these conditions rank errors above chance yet far from perfectly (0.82--0.87), vacuity offers no advantage over the detection score, and with ADFM vacuity is at chance (0.49).
 
The exploratory crop-level control points the same way. Re-evaluated on isolated detection crops with the same frozen weights, vacuity ranks errors worse than over all detections (UAUC 0.71 and 0.60 versus 0.96 and 0.97; Section~\ref{sec:results}), so isolating the object does not yield a better error signal than the detection pipeline already provides.

\vspace{-0.4em}
 
\subsection{Why Dempster-Shafer Fusion's Advantages Did Not Translate}
 
\looseness=-1 DS fusion is a principled method, but its theoretical advantage did not translate into empirical gains here. That advantage is conditional on conflict that distinguishes a reliable modality from an unreliable one; absent such conflict, DS has no mechanism to outperform a plain average. The reason is structural: Dempster's rule treats agreement between modalities as independent corroboration and sharpens concordant evidence rather than averaging it. Two mechanisms are observed directly, and a third is plausible but untested. First, redundancy: where modalities observe the same scene --- E2 under clean conditions, RGB and IR throughout E3 --- evidence accumulation has little to arbitrate, and the parameter-free average captures the available benefit without the combination rule's overhead. Second, the conflict mechanism never activates: in E2, inter-modal conflict $C$ approaches but stays below the fallback threshold under standard noise and only marginally exceeds it under the all-noise foil, where the fallback fires without rescuing the collapse; in E3, with a single focal class, $C$ is structurally zero, and the disagreement proxy $D$ correlates only weakly with the fused uncertainty, so the uncertainty-driven down-weighting that motivates DS does not occur. Third, in E3 the decision-level fusion of spatially-registered detections may impose a localisation and duplication penalty: both fusion rules stay below the stronger modality in MOTA and roughly double the identity switches of the RGB baseline (291 and 270 versus 144), but the duplication mechanism was not tested. The single-class degeneracy runs deeper than the conflict term: with one focal class the naive EDL objective is itself identically zero (Section~\ref{sec:loss}), so both formalisms --- DS conflict and the evidential loss --- require an explicit background hypothesis to be non-trivial in single-class detection.

\vspace{-0.4em}
 
\subsection{Practical Implications}
 
\looseness=-1 These results suggest a default ordering of methods for the redundant-sensor, imbalanced-detection systems studied here. At the classification level, evidential heads trained on frozen backbones improved accuracy at no inference cost (E2); in tracking (E1) they did not. For fusing redundant modalities, parameter-free averaging matches DS combination in accuracy at lower complexity and should be the default; DS earns its overhead only where modalities are genuinely complementary and carry a conflict signal with dynamic range. Where the practical concern is calibrated confidence, temperature scaling reached the calibration of the best evidential conditions at lower cost (E2). And in dense detection regimes, no predictive-uncertainty signal --- evidential vacuity or detection confidence --- should be used to gate sensors or filter false positives without prior validation on the target distribution: a high aggregate AUROC, dominated by easy low-score errors, does not show that a signal separates confident false positives.
 
\subsection{Relation to Prior Work}
 
\looseness=-1 Li et al.~\cite{li_stabilizing_2024} and Shao et al.~\cite{shao_dual-level_2024, shao_ua-fusion_2025} demonstrated that DS and uncertainty-aware fusion stabilise multispectral pedestrian detection and 3D object detection respectively. Most directly, Suttorp et al.~\cite{suttorp_uncertainty_2026} report a closely related null result in RGB--thermal anti-UAV perception: multimodal evidential fusion improves over either single stream, but conflict-discounted belief fusion is empirically indistinguishable from undiscounted averaging because the benchmark produces almost no inter-modal conflict. The present results extend that boundary across different sensor combinations and task levels: in the settings examined here --- redundant modality pairs, extreme detection-level class imbalance, and, for E3, a single focal class that nulls the conflict term --- the conditions evidence-combination rules exploit are absent, so DS confers no advantage over averaging. The GMU comparison fits this picture: the reproduced GMU collapses to near-majority accuracy (Section~\ref{sec:results}). The gate's softmax weights collapsed to [Audio$=$0.00, Video$=$1.00, RF$=$0.00] on both clean and noisy data: the GMU reached 100\% on clean inputs, so the gate never learned to fall back when video degrades at test time.
 
\subsection{Validity and Reliability}
 
\paragraph{Reproducibility.} The E2 and E3 baselines were not reproducible as published, and E1 deviates in its evaluation protocol; the documented repairs (Appendix~\ref{sec:apx:repro_dev}) are a prerequisite for any valid measurement. Because each baseline was repaired toward its published behaviour before the evidential components were introduced, the negative results cannot be attributed to weakened baselines --- a controlled retrained baseline is the comparator in every ablation (for E3, Stage~2 of the baselines is retrained under the same protocol as the evidential conditions).
 
\paragraph{Statistical reliability.} The effective sample size is bounded by independent clips or sequences, not by segments or frames. The E2 clip-clustered bootstrap ($n=33$) widens confidence intervals three- to six-fold relative to naive segment-level intervals; both evidential gains over Late Fusion (with averaging and with DS) survive this correction. E1 evaluates 50 validation sequences, and E3 reports single-seed tracking, so the small DS-versus-averaging MOTA difference in E3 (0.65 points) is not claimed in either direction. The quantitative margins are thus setting-specific.
 
\paragraph{Generalisability and scalability.} Degradation is synthetic throughout (additive noise on audio; flip, rotation, colour jitter, blur and salt-and-pepper noise on video; spectrogram-domain rather than I/Q-level perturbation for RF), E3 fuses only two modalities, and results are specific to the three benchmarks. The negative findings on DS fusion and gating are expected to generalise to other redundant-modality, imbalanced-detection settings, since they follow from structural properties (redundancy, class imbalance, a single focal class) rather than from tuning; the E2 accuracy gain of the evidential heads is consistent with the broader evidential-learning literature~\cite{sensoy_evidential_2018, gao_comprehensive_2025}.
 
\subsection{Ethical Considerations}
 
Counter-UAS is a dual-use defence technology: the same detection-fusion pipeline that protects civilian airspace can be directed at any airborne target, and the study's framing around airport and conflict-zone incursions should not obscure that. Two findings carry specific ethical weight. First, the evaluation of uncertainty signals is safety-relevant. An operator who trusted evidential vacuity to flag false positives --- the use case the framework set out to enable --- would be relying on an aggregate AUROC dominated by easy low-score errors; among confident detections, where targeting decisions are made, vacuity ranked errors no better than the detection score. Such signals need validation on the target distribution before use inside a high-stakes decision loop. Second, the benchmarks are documented by their creators, but consent for individuals incidentally captured on camera is not; accordingly, all conclusions are reported under reproduced conditions on public data, and no claim of field-validated robustness is made.

\section{Conclusion}
\label{sec:conclusion}
This study asked whether EDL heads, DS evidence fusion, and uncertainty-driven gating improve detection accuracy, multi-modal robustness, and computational efficiency in anti-UAV systems. The controlled ablation across three benchmarks returns an asymmetric answer: evidential heads improve classification accuracy but not tracking accuracy (E1), and their uncertainty adds no usable signal beyond predictive entropy or the detection score; DS fusion fails to improve multi-modal robustness over simple averaging, and uncertainty-driven gating yields no computational-efficiency gain.

Dempster--Shafer (DS) fusion is theoretically principled, but it did not significantly outperform probability averaging in the evaluated settings. In E2, DS fusion achieved 97.53\% accuracy compared with 97.20\% for probability averaging (one-sided $p=0.332$), and in E3 the two differed by less than one RGB MOTA point (56.57 vs.\ 55.91, single seed). This lack of improvement is consistent with the structure of the data: the modalities are largely redundant, conflict exhibits little dynamic range in E2, and conflict is structurally zero in the single-class setting of E3. Under such conditions, DS fusion has little opportunity to exploit inter-modal disagreement and performs no better than simple averaging.

The evidential formulation produced clearer benefits in classification than in detection. In E2, evidential heads trained on the frozen backbones improved accuracy by 4.8 percentage points, with the gain remaining robust under a clip-clustered bootstrap analysis; their entropy also ranked classification errors better than the sigmoid baseline, although this difference was not significant at the clip level. In E1, replacing the detector's objectness and classification terms did not improve tracking accuracy (0.562 vs.\ 0.572) and roughly tripled the tracker-on-absent rate, which here means more boxes on target-absent frames. These results did not translate into reliable false-positive reduction through epistemic uncertainty. At the detection level, vacuity ranks errors about as well as the detection score over all detections, a figure dominated by easy low-score errors, and nominally worse among confident detections. An exploratory crop-level control shows that isolating detected objects lowers this ranking.

Per-modality uncertainty also proved unsuitable as a practical control signal for sensor management. In E2, gating preserved accuracy only when the mechanism was nearly inactive and reduced accuracy by 5.7 percentage points at the selected operating point. In E3, the gating mechanism did not engage at all. Moreover, no latency reduction was realised because the benchmark evaluates every modality encoder before the gating decision is made. These findings indicate that, although evidential heads can improve classification accuracy, the evaluated uncertainty signals are not sufficiently reliable or operationally informative to support either improved fusion or efficient sensor selection.

These answers are qualified by the study's limitations: degradation was synthetic, E1 was evaluated on validation rather than withheld test sequences, E2's effective sample size is 33 independent clips, and E3 reports single-seed tracking with a two-modality fusion. The quantitative margins are therefore setting-specific.

Relative to the gap identified in Section~\ref{sec:related_work}, the central contribution is to characterise where uncertainty-aware methods help and where they do not, and to separate the effect of training evidential heads from the evidential uncertainty they are usually packaged with: the heads improve classification accuracy, while their vacuity adds nothing beyond predictive entropy in the primary classification conditions and ranks detection errors about as well as the detection score overall but less well among confident detections. Future work should test imbalance-aware evidential formulations (e.g.\ heatmap-level uncertainty), evaluate uncertainty signals specifically on confident false positives, and evaluate DS on sensor pairs with genuine inter-modal conflict.

\bibliographystyle{ACM-Reference-Format}
\bibliography{bibtex_acm_updated}

\appendix
\clearpage
\onecolumn

\section{Training and Inference Protocols}
\label{sec:apx:protocols}

\setcounter{table}{0}
\renewcommand{\thetable}{A.\arabic{table}}

\begin{table*}[htbp]
\centering
\caption{Baseline selection. Each baseline confines evidential reasoning (if present) to downstream modules, leaving detection heads with standard sigmoid outputs. This is the structural gap EviDS-UAV is designed to investigate.}
\label{tab:baseline_selection}
\small
\begin{tabular}{p{2.0cm} p{3.0cm} p{2.5cm} p{8cm}}
\toprule
\textbf{Experiment} & \textbf{Baseline} & \textbf{Dataset} & \textbf{Selection rationale} \\
\midrule
E1 & EDTC~\cite{zhu_evidential_2023} & AntiUAV600 & Developed by the AntiUAV600 authors; foundational detection-tracking collaboration paradigm. Ranks 2nd on the AntiUAV600 leaderboard behind ADTC~\cite{liu_learning_2025}, which extends EDTC without modifying the evidential component; both outperform standard trackers by $>$20~pp Acc. Neither applies EDL to the detector. \\
\addlinespace
E2 & TRIDENT~\cite{alla_trident_2025} & TRIDENT & The only published tri-modal (RGB + audio + RF) drone classifier. \emph{De facto} baseline by construction: no competing method exists on this dataset. While GMU fusion is reproduced, it collapses (serving as a cautionary baseline); thus, Late Fusion (92.42\%) serves as the primary comparator. Both lack per-sensor reliability modelling. \\
\addlinespace
E3 & MMA-SORT~\cite{xu_tri-modal_2025} & MM-UAV & Baseline for the MM-UAV benchmark, reproducing the published ADFM channel-attention fusion. ADFM's weighting carries no per-stream reliability signal, so a degraded stream (IR under thermal crossover, RGB in low light or smoke) keeps an attention weight uncorrelated with its reliability --- the structural gap EviDS-UAV targets. MM-UAV's small targets and high RGB/IR redundancy extend the imbalance and redundancy themes examined in E1 and E2. \\
\bottomrule
\end{tabular}
\end{table*}

\begin{table*}[htbp]
\centering
\caption{E1 training and inference protocol. Only the YOLOv5s detector is retrained; the CvT tracker uses the released Stage-2 checkpoint throughout.}
\label{tab:e1_protocol}
\small
\begin{tabular}{p{2.5cm} p{5.5cm} p{5.5cm}}
\toprule
& \textbf{Training} & \textbf{Inference} \\
\midrule
\textbf{Input} & Thermal IR frames with GT bounding boxes from 270 training sequences & Thermal IR video sequences (50 val sequences) \\
\textbf{Model} & YOLOv5s (sigmoid or EDL head variant) & Full EDTC pipeline: YOLOv5s detector + CvT tracker \\
\textbf{Output} & Trained detector checkpoint & Per-frame bounding box predictions + confidence/uncertainty scores \\
\textbf{Splits} & 270 train / 30 tune (fixed seed 42) & 50 official validation sequences (held-out) \\
\textbf{Epochs} & 20 (matching Zhu et al.~\cite{zhu_evidential_2023}) & --- \\
\textbf{GT usage} & Bounding box supervision for detector training & IoU-based Acc computation; per-frame presence flags for absence penalty \\
\textbf{Checkpoints} & Best YOLOv5 fitness ($0.1\,\text{mAP}_{50} + 0.9\,\text{mAP}_{50:95}$) on 30 tune sequences & Released CvT tracker (\texttt{best.pt}); detector checkpoint selected per condition \\
\bottomrule
\end{tabular}
\end{table*}

\begin{table*}[htbp]
\centering
\caption{E2 training and inference protocol. Unimodal models are trained first; fusion models train only the fusion head with frozen backbones. Evaluation operates at the segment level (0.25\,s intervals).}
\label{tab:e2_protocol}
\small
\begin{tabular}{p{2.5cm} p{5.5cm} p{5.5cm}}
\toprule
& \textbf{Training} & \textbf{Inference} \\
\midrule
\textbf{Input} & Per-modality features from 212 training clips (clean, no augmentation), segmented into 0.25\,s intervals & Per-modality features from 33 test clips (1,320 segments), with noise applied at test time to audio and video \\
\textbf{Stage 1} & Train 5 unimodal models independently (Table~\ref{tab:e2_arch}) & --- \\
\textbf{Stage 2} & Train the fusion head (GMU, late fusion) or, for the evidential conditions, one evidential head per modality with $\mathcal{L}_{\text{EDL}}$ (KL annealing over 500 optimiser steps, $\approx$6.6 of 30 epochs), all with frozen Stage-1 backbones; DS and averaging are parameter-free & Full pipeline: frozen backbones feed the heads, then fusion, then a per-segment binary prediction \\
\textbf{Output} & Unimodal + fusion checkpoints & Per-segment binary prediction + per-modality uncertainty scores \\
\textbf{Noise} & Not applied during training (trained on clean data) & Additive noise to audio (librosa); random flip, rotation, colour jitter, blur, and salt-and-pepper noise on video; RF evaluated clean only (see Section~\ref{sec:apx:repro_dev}) \\
\textbf{GT usage} & Binary drone/background labels for BCE or $\mathcal{L}_{\text{EDL}}$ & Segment-level accuracy, macro-F1, ECE \\
\bottomrule
\end{tabular}
\end{table*}

\begin{table*}[htbp]
\centering
\caption{E3 training and inference protocol. Training follows a two-stage procedure: Stage~1 pretrains the full dual-stream YOLOX-s backbone with sigmoid heads; Stage~2 fine-tunes the detection heads (and ADFM where it is retained, conditions~a--c) with the backbone, FPN, and OGAA alignment frozen; DS and averaging fusion are parameter-free. Inference uses the full MMA-SORT pipeline with event-based identity association.}
\label{tab:e3_protocol}
\small
\begin{tabular}{p{2.5cm} p{5.5cm} p{5.5cm}}
\toprule
& \textbf{Training} & \textbf{Inference} \\
\midrule
\textbf{Input} & Paired RGB and IR frames with COCO-format bounding box annotations from 1,080 training sequences; no data augmentation (no Mosaic, no mixup, no random resizing) & Paired RGB and IR video from 121 test sequences; event camera data for tracking association only \\
\textbf{Stage 1} & Train dual-stream YOLOX-s (backbone + FPN + sigmoid heads) for 100 epochs; AdamW with betas (0.9, 0.999) & --- \\
\textbf{Stage 2} & Fine-tune the detection heads (sigmoid or evidential) for 50 epochs (40 for STN, as in the authors' released configuration) with the backbone, FPN, and OGAA alignment frozen; ADFM is trained only where retained (conditions~a--c), while DS and averaging are parameter-free. Evidential heads use $\mathcal{L}_{\text{EDL}}$ with KL annealing over 10 epochs & Full MMA-SORT pipeline: dual-stream YOLOX-s detector + tracking with event-based identity association; detection threshold conf = 0.3 (matching published \texttt{final\_def.sh}) \\
\textbf{Output} & Trained Stage-2 detector checkpoint (best AP$_{50}$ on 120 tune sequences) & Per-frame bounding box predictions, per-modality confidence/uncertainty scores, and tracking trajectories \\
\textbf{Splits} & 1,080 train / 120 tune / 121 test (tune split = 10\% of official 1,200 training sequences; fixed seed 42) & 121 official test sequences (MM-UAV toolkit evaluation) \\
\textbf{GT usage} & Bounding box supervision for regression; binary presence labels for $\mathcal{L}_{\text{EDL}}$ or sigmoid BCE & MOTA, HOTA, IDF1, IDs via MM-UAV toolkit; detection-level ECE and UAUC as in E1--E2 \\
\textbf{Checkpoints} & Stage-1 weights loaded from \texttt{yolox\_s\_2\_stream} (published) and frozen; Stage-2 checkpoint selected per condition by AP$_{50}$ on the tune split (selected epochs: DefConv 18/50, STN 39/40, ADFM 32/50, Average 12/50, DS 14/50) & Retrained Stage-2 weights for all conditions; (f) uses the checkpoint of (e), (g) that of (a) with $T$ fitted on the tune split \\
\bottomrule
\end{tabular}
\end{table*}

\begin{table*}[htbp]
\centering
\caption{TRIDENT unimodal architectures (E2). Each modality is trained independently; backbone weights are frozen during fusion training. The RF backbone uses 3D convolutions architecturally, but operates with a temporal depth of 1 frame per 0.25\,s segment (4\,FPS $\times$ 0.25\,s\,$=$\,1).}
\label{tab:e2_arch}
\footnotesize
\setlength{\tabcolsep}{6pt}
\begin{tabular}{@{} l l l l @{}}
\toprule
\textbf{Modality} & \textbf{Segment input} & \textbf{Backbone(s)} & \textbf{Head} \\
\midrule
RGB Video & 7 frames $\times$ 112$\times$112 RGB & Inflated 3D ResNet-10 / MobileNet & Sigmoid \\
\addlinespace
Audio & 0.25\,s WAV $\to$ MFCC ($1 \times 40 \times 40$) & LeNet / VGG-19 & Sigmoid \\
\addlinespace
RF & 1 spectrogram $\times$ 112$\times$112 & Inflated 3D ResNet-10\textsuperscript{$\dagger$} & Sigmoid \\
\bottomrule
\multicolumn{4}{l}{\vspace{2pt}\scriptsize \textsuperscript{$\dagger$}Temporal depth$\,=\,$1; the 3D architecture reduces to 2D convolution in practice.} \\
\multicolumn{4}{l}{\scriptsize RF MobileNet variant omitted (5 unimodal models trained, not 6).} \\
\end{tabular}
\end{table*}

\clearpage

\section{Ablation Configurations}
\label{sec:apx:ablations}

\setcounter{table}{0}
\renewcommand{\thetable}{B.\arabic{table}}

\begin{table*}[htbp]
\centering
\caption{E1 ablation conditions. All conditions use the same 270/30/50 data split, the same CvT tracker checkpoint, and the same evaluation protocol.}
\label{tab:e1_ablation}
\small
\begin{tabular}{c l p{8cm}}
\toprule
\textbf{Cond.} & \textbf{Label} & \textbf{Description} \\
\midrule
(a) & Sigmoid-270 & YOLOv5s with sigmoid head, retrained on 270 sequences for 20 epochs. Fair ablation baseline: isolates the effect of the evidential head from data/training differences relative to the released checkpoint. \\
(b) & EDL-270 & YOLOv5s with evidential Dirichlet head, trained on 270 sequences with $\mathcal{L}_{\text{EDL}}$. Core comparison: same architecture, data, and epochs as (a), differing only in the objectness and classification terms, which are replaced by a two-class (background/UAV) $\mathcal{L}_{\text{EDL}}$. \\
(c) & EDL-270 + $\tau_{\text{det}}$ & As (b), with detection-level uncertainty thresholding: detections with $u > \tau_{\text{det}}$ are suppressed before entering the tracker. Tests whether detector-level uncertainty filtering reduces false-positive-initiated tracks. \\
(d) & Sigmoid-270 + TempScale & As (a), with post-hoc temperature scaling applied to sigmoid outputs. On 98.9\% of detector-active frames there is no detection (confidence 0), so NLL decreases monotonically with $T$ and has no finite minimiser; $T=200$ is used as a practical cap, which maps all confidences to $\approx 0.5$. Calibration control: temperature scaling lowers Det.\ ECE only through this compression and, being monotonic, leaves switching (TrkOnAbs) unchanged. \\
\bottomrule
\end{tabular}
\end{table*}

\begin{table*}[htbp]
\centering
\caption{E2 ablation conditions. All conditions use the same corrected codebase (Section~\ref{sec:apx:repro_dev}), the same frozen unimodal backbones, and identical noise protocols. RF is evaluated clean-only except where stated. Conditions (d$'$) and (g) share identical per-modality noise (audio, video, and spectrogram-domain RF at 20\,dB SNR); they differ only in the fusion rule and thus form a controlled foil pair.}
\label{tab:e2_ablation}
\small
\begin{tabular}{c l p{8cm}}
\toprule
\textbf{Cond.} & \textbf{Label} & \textbf{Description} \\
\midrule
(a) & Unimodal baselines & Five individual sigmoid models (Table~\ref{tab:e2_arch}), evaluated independently. Establishes per-modality performance floor. \\
(b) & Late Fusion (sigmoid) & Learned weighted combination of three sigmoid outputs. Reproduced baseline. \\
(c) & GMU (sigmoid) & Gated Multimodal Unit with softmax gating (sigmoid output). Reproduced baseline. \\
(d) & Evidential + Average & Evidential heads on all modalities; fusion by simple averaging of predicted class probabilities $\hat{p}_k$. Isolates the effect of evidential heads from the DS fusion rule. \\
(d') & Evidential + Average + Noise & As (d), with synthetic noise injected across all modalities simultaneously to establish a degraded fusion baseline without DS reasoning. \\
(e) & Evidential + DS Fusion & Evidential heads with iterative DS combination (Section~\ref{sec:ds_fusion}). Core comparison against (b) and (c). \\
(f) & Evidential + DS + Gating & Full EviDS-UAV: DS fusion with temporal sensor gating (Section~\ref{sec:temporal_gating}). Tests the accuracy--efficiency trade-off. Hysteresis $\tau_{\text{high}}=0.35$, $\tau_{\text{low}}=0.05$, $N=10$, $\tau_{\text{fused}}=0.03$, tuned on the validation split.\\
(g) & Evidential + DS + RF noise & As (e), with spectrogram-domain noise added to RF (time-frequency masking + additive pink noise). Tests DS fusion under tri-modal degradation. This is not I/Q-faithful but provides a controlled degradation absent from (a)--(f) except (d$'$), where RF is evaluated clean-only. \\
\bottomrule
\end{tabular}
\end{table*}

\begin{table*}[htbp]
\centering
\caption{E3 ablation conditions. All conditions use a 1080/120/121 (train/tune/test) data split derived from the official 1,200 training and 121 test sequences (10\% of training held out as tune split), the same dual-stream YOLOX backbone, and identical evaluation protocol (MM-UAV toolkit, MOTA/HOTA/IDF1/IDs + detection ECE). OGAA spatial alignment is retained in all conditions. Stage 2 is retrained for all conditions (backbone + FPN frozen from the published Stage-1 weights), with checkpoint selection on the tune split; (f) uses the checkpoint of (e) and (g) that of (a).}
\label{tab:e3_ablation}
\small
\begin{tabular}{c l p{10cm}}
\toprule
\textbf{Cond.} & \textbf{Label} & \textbf{Description} \\
\midrule
(a) & MMA-SORT DefConv & OGAA (deformable convolution) + ADFM channel-attention fusion + sigmoid heads. Baseline retrained under the same protocol as (c)--(e): ablation comparator. \\
\addlinespace
(b) & MMA-SORT STN & OGAA (spatial transformer) + ADFM + sigmoid heads, 40 Stage-2 epochs as in the authors' configuration. Second alignment variant; tests whether alignment choice interacts with evidential components. \\
\addlinespace
(c) & Evidential + ADFM & Replace sigmoid heads with evidential Dirichlet heads on both YOLOX streams. ADFM feature-level channel-attention fusion retained (Fusion0/1/2 unchanged). Each stream produces per-modality Dirichlet outputs ($\alpha, u$). ADFM fuses features before the heads; the heads receive aligned+enhanced features and output per-modality evidence. Isolates EDL head effect from fusion changes. \\
\addlinespace
(d) & Evidential + Average & As (c), but replace ADFM channel-attention with \emph{decision-level probability averaging}: each YOLOX stream carries a full independent evidential head, and the per-anchor class probabilities $\hat{p}_k=\alpha_k/S$ from the two heads are averaged before the confidence threshold, with RGB bounding-box geometry retained. Isolates the DS fusion rule: if (e) outperforms (d), the gain is attributable to DS evidence combination rather than to evidential heads alone. Mirrors E2 condition~(d). \\
\addlinespace
(e) & Evidential + DS Fusion & Core EviDS-UAV for E3. Evidential heads on both streams + DS evidence fusion at decision level. ADFM removed; OGAA spatial alignment retained. Fusion operates on per-stream Dirichlet outputs ($\alpha_{\text{rgb}}, u_{\text{rgb}}$) and ($\alpha_{\text{ir}}, u_{\text{ir}}$) after the detection heads, producing fused belief masses and fused uncertainty. Event-based tracking association unchanged. \\
\addlinespace
(f) & Evidential + DS + Gating & Full EviDS-UAV. As (e), with temporal sensor gating between RGB and IR streams (hysteresis thresholds $\tau_{\text{low}}, \tau_{\text{high}}$; suppression window $N$). Constraint: at least one modality always active. Tests accuracy--efficiency trade-off. \\
\addlinespace
(g) & Baseline + TempScale & Post-hoc temperature scaling on (a) sigmoid outputs. $T$ fitted by NLL minimisation on the held-out tune split. Calibration control: if (c) or (e) achieves comparable ECE to (g), EDL provides well-calibrated uncertainty without post-hoc correction. \\
\bottomrule
\end{tabular}
\end{table*}

\clearpage

\section{Full Results Tables}
\label{sec:apx:full_results}

\setcounter{table}{0}
\renewcommand{\thetable}{C.\arabic{table}}

\begin{table*}[htbp]
\centering
\caption{Full E1 results on 50 AntiUAV600 validation sequences (56,301 frames; 1,735 absent); complete version of the focused Table~\ref{tab:e1_results}. Acc is IoU-based accuracy with absence penalty (Section~\ref{sec:e1}). Det.\ ECE and Trk.\ ECE are computed over detector-active and tracker-active frames respectively. TrkOnAbs = fraction of absent frames where the tracker remains in control (uncertainty $< \theta_{eh}$) without delegating to the detector; AbsBox = number of target-absent frames on which a box is output. Det.\ ECE is dominated by detector-active frames without a detection (confidence 0) and mainly reflects the miss rate.}
\label{tab:e1_results_full}
\begin{tabular}{l c c c c c c c}
\toprule
\textbf{Condition} & \textbf{Acc} $\uparrow$ & \textbf{Det.\ ECE} $\downarrow$ & \textbf{Trk.\ ECE} $\downarrow$ & \textbf{Overall ECE} $\downarrow$ & \textbf{TrkOnAbs} & \textbf{AbsBox} $\downarrow$ & \textbf{FPS} \\
\midrule
Released EDTC (\texttt{best.pt}) & 0.576 & 0.877 & 0.040 & 0.246 & 8.6\% & 132 & $\sim$20 \\
(a) Sigmoid-270 & \textbf{0.572} & 0.866 & 0.041 & 0.232 & 5.5\% & \textbf{80} & $\sim$20 \\
(b) EDL-270 & 0.562 & 0.875 & \textbf{0.036} & 0.218 & 17.4\% & 290 & $\sim$20 \\
(c) EDL-270 + $\tau_{\text{det}}\!=\!0.5$ & 0.547 & 0.888 & 0.040 & 0.268 & 7.3\% & 117 & $\sim$20 \\
(d) Sigmoid-270 + TempScale ($T\!=\!200$) & \textbf{0.572} & \textbf{0.394} & 0.041 & \textbf{0.123}$^\dagger$ & 5.5\% & \textbf{80} & $\sim$20 \\
\bottomrule
\end{tabular}%
\vspace{4pt}
\begin{minipage}{\textwidth}
\footnotesize
$^\dagger$Overall ECE for (d) is approximated by frame-count weighting of the temp-scaled detector confidences with unchanged tracker confidences. Bold: best per column among conditions (a)--(d); the released checkpoint row is reference only. Throughput is $\sim$20\,FPS for all conditions (shared YOLOv5s + CvT-21 architecture).
\end{minipage}
\end{table*}

\begin{table*}[htbp]
\centering
\caption{Full E2 results on 1,320 TRIDENT test segments (33 clips $\times$ 40 segments); complete version of the focused Table~\ref{tab:e2_results}, adding the five unimodal backbones. Mean $\pm$ sample standard deviation across noise seeds 42/123/456; ECE is top-label (15 bins). RF is evaluated clean-only throughout. UAUC measures how well uncertainty ranks classification errors --- 0.5 is random, 1.0 is perfect. Predictive entropy ($\text{UAUC}_{\text{ent}}$) is computed identically for all models (from $\hat{p}$ for evidential heads); vacuity ($\text{UAUC}_{\text{vac}}$) is additionally reported for the evidential conditions only.}
\label{tab:e2_results_full}
\begin{tabular}{l l c c c c c}
\toprule
\textbf{Condition} & \textbf{Noise} & \textbf{Acc (\%)} $\uparrow$ & \textbf{Macro-F1} $\uparrow$ & \textbf{ECE} $\downarrow$ & \textbf{UAUC}$_{\text{vac}}$ $\uparrow$ & \textbf{UAUC}$_{\text{ent}}$ $\uparrow$ \\
\midrule
\multicolumn{7}{l}{\textit{Unimodal baselines (a)}} \\
\quad LeNet (Audio) & Noisy & 65.30 $\pm$ 0.40 & 0.4547 $\pm$ 0.0085 & 0.3228 $\pm$ 0.0021 & --- & 0.718 \\
\quad VGG (Audio) & Noisy & 60.58 $\pm$ 0.24 & 0.5175 $\pm$ 0.0049 & 0.3577 $\pm$ 0.0040 & --- & 0.455 \\
\quad ResNet-10 (Video) & Noisy & 36.36 $\pm$ 0.00 & 0.2667 $\pm$ 0.0000 & 0.6359 $\pm$ 0.0000 & --- & 0.730$^\dagger$ \\
\quad MobileNet (Video) & Noisy & 69.70 $\pm$ 0.00 & 0.6967 $\pm$ 0.0000 & 0.3207 $\pm$ 0.0004 & --- & 0.158 \\
\quad ResNet-10 (RF) & Clean & 81.74 $\pm$ 0.00 & 0.8159 $\pm$ 0.0000 & 0.1491 $\pm$ 0.0000 & --- & 0.865 \\
\midrule
\multicolumn{7}{l}{\textit{Fusion baselines}} \\
\quad (b) Late Fusion & Noisy & 92.42 $\pm$ 0.20 & 0.920 $\pm$ 0.002 & 0.154 $\pm$ 0.001 & --- & 0.496 \\
\quad (c) GMU & Noisy & 61.44 $\pm$ 0.47 & 0.416 $\pm$ 0.009 & 0.317 $\pm$ 0.004 & --- & 0.506 \\
\midrule
\multicolumn{7}{l}{\textit{EviDS-UAV conditions}} \\
\quad (d) EDL + Average & Noisy & 97.20 $\pm$ 0.08 & 0.970 $\pm$ 0.001 & 0.254 $\pm$ 0.001 & 0.880 & \textbf{0.942} \\
\quad (e) EDL + DS & Noisy & \textbf{97.53} $\pm$ 0.17 & \textbf{0.973} $\pm$ 0.002 & 0.091 $\pm$ 0.003 & \textbf{0.935} & \textbf{0.942} \\
\quad (f) EDL + DS + Gating & Noisy & 91.82 $\pm$ 1.55 & 0.911 $\pm$ 0.018 & \textbf{0.054} $\pm$ 0.007 & 0.869 & 0.813 \\
\midrule
\multicolumn{7}{l}{\textit{Full-degradation foil}} \\
\quad (d$'$) EDL + Average & Noisy (all) & 65.73 $\pm$ 0.27 & 0.449 $\pm$ 0.009 & 0.148 $\pm$ 0.004 & 0.726 & 0.727 \\
\quad (g) EDL + DS$^\ddagger$ & Noisy (all) & 63.91 $\pm$ 0.04 & 0.397 $\pm$ 0.001 & 0.328 $\pm$ 0.002 & 0.812 & 0.778 \\
\bottomrule
\multicolumn{7}{l}{\footnotesize $^\dagger$ResNet-10 (Video) UAUC is inflated by a constant-prediction artefact: the model predicts Drone 0\% of the time (collapsed to the minority class),} \\
\multicolumn{7}{l}{\footnotesize \quad so proximity to the 0.5 decision boundary correlates with class membership, not error likelihood.} \\
\multicolumn{7}{l}{\footnotesize $^\ddagger$(g) collapses to majority-class prediction (Drone on 99.7--99.8\% of segments); its UAUC is a constant-prediction artefact of the} \\
\multicolumn{7}{l}{\footnotesize \quad same kind as ResNet-10 (Video)'s, though in the opposite direction (majority- rather than minority-class collapse).} \\
\multicolumn{7}{l}{\footnotesize Bold: best per column among fusion conditions (b)--(g); unimodal rows and artefact-flagged values excluded.} \\
\end{tabular}%
\end{table*}

\begin{table*}[htbp]
\centering
\caption{Full E3 results on MM-UAV (121 test sequences); complete version of the focused Table~\ref{tab:e3_results}. ECE and UAUC are detection-level (RGB stream, post-NMS, IoU $\geq$ 0.5 to GT). Conditions (a)--(b) are the baselines retrained under the same protocol; (c)--(f) are evidential ablations and (g) a calibration control; (c$'$) is a diagnostic row at a lowered confidence threshold. The threshold conf$=$0.3 matches the published \texttt{final\_def.sh} operating point: applied per-stream in (c)/(c$'$) and post-fusion in (d)/(e)/(f) (Section~\ref{sec:e3}, Table~\ref{tab:e3_confsweep}). Single training seed per condition. Bold: best per column among operating conditions (a), (b), (d)--(g); collapsed diagnostic rows are unmarked.}
\label{tab:e3_results_full}
\resizebox{\textwidth}{!}{%
\begin{tabular}{l cccc cccc c c c}
\toprule
& \multicolumn{4}{c}{RGB} & \multicolumn{4}{c}{IR} & & & \\
\cmidrule(lr){2-5} \cmidrule(lr){6-9}
\textbf{Condition} & MOTA $\uparrow$ & HOTA $\uparrow$ & IDF1 $\uparrow$ & IDs $\downarrow$ & MOTA $\uparrow$ & HOTA $\uparrow$ & IDF1 $\uparrow$ & IDs $\downarrow$ & ECE $\downarrow$ & UAUC$_{\text{vac}}$ $\uparrow$ & UAUC$_{1-\text{conf}}$ $\uparrow$ \\
\midrule
(a) MMA-SORT DefConv          & \textbf{64.28} & 56.72 & 74.98 & \textbf{144} & \textbf{80.74} & \textbf{67.09} & \textbf{84.32} & \textbf{140} & 0.051 & --- & 0.959 \\
(b) MMA-SORT STN              & 64.22 & \textbf{57.07} & \textbf{75.97} & \textbf{144} & 79.09 & 66.04 & 82.83 & 187 & 0.052 & --- & \textbf{0.964} \\
\midrule
(c) Evidential + ADFM (conf=0.3) & 0.02  & 0.70  & 0.15  & 0   & 1.22  & 5.04  & 2.81  & 2   & 0.475 & 0.817 & 0.893 \\
(c$'$) Evidential + ADFM (conf=0.1) & 0.14 & 1.58 & 0.36 & 0 & 1.87 & 7.19 & 4.51 & 26 & 0.475$^\ddagger$ & 0.817$^\ddagger$ & 0.893$^\ddagger$ \\
\midrule
(d) Evidential + Average      & 55.91 & 54.79 & 71.11 & 291 & ---$^\dagger$ & ---$^\dagger$ & ---$^\dagger$ & ---$^\dagger$ & \textbf{0.035} & 0.960 & 0.957 \\
(e) Evidential + DS Fusion    & 56.57 & 55.47 & 72.35 & 270 & ---$^\dagger$ & ---$^\dagger$ & ---$^\dagger$ & ---$^\dagger$ & 0.036 & \textbf{0.972} & 0.959 \\
(f) Evidential + DS + Gating  & 56.57 & 55.47 & 72.35 & 270 & ---$^\dagger$ & ---$^\dagger$ & ---$^\dagger$ & ---$^\dagger$ & 0.036$^\S$ & \textbf{0.972}$^\S$ & 0.959$^\S$ \\
\midrule
(g) Baseline + TempScale ($T\!=\!0.626$) & 62.16 & 57.00 & 74.89 & 183 & 79.88 & 66.72 & 83.38 & 170 & 0.058 & --- & 0.944 \\
\bottomrule
\end{tabular}%
}
\vspace{4pt}
\begin{minipage}{\textwidth}
\footnotesize
$^\dagger$Conditions (d), (e), and (f) return a single fused detection in RGB coordinates (Sec.~\ref{sec:e3}); per-stream IR metrics are structurally undefined.\\
$^\ddagger$(c$'$) uses the same model checkpoint as (c) and lowers only the detection threshold (new tracks still require a score above 0.7); ECE and UAUC are identical to (c), computed on the 12{,}955 annotated frames of the 121 test sequences, independent of the tracker threshold.\\
$^\S$(f) uses the same model checkpoint as (e); ECE, UAUC$_{\text{vac}}$, and UAUC$_{1-\text{conf}}$ are identical to (e); its tracker output is identical to that of (e).
\end{minipage}
\end{table*}

\clearpage

\section{Baseline Reproduction Deviations}
\label{sec:apx:repro_dev}

\setcounter{table}{0}
\renewcommand{\thetable}{D.\arabic{table}}

\begin{table*}[htbp]
\centering
\caption{Consolidated reproduction deviations from published baselines.}
\label{tab:deviations}
\small
\begin{tabular}{c l p{9cm}}
\toprule
\textbf{Exp.} & \textbf{Deviation} & \textbf{Reason and consequence} \\
\midrule
\multirow{3}{*}{E1} & Evaluation on 50 val sequences (not 250 test) & Competition evaluation server unavailable; results not directly comparable to published leaderboard figures. \\
\cmidrule{2-3}
& Acc formula uses $\Sigma q_t / T$ (all frames) & Matches the released evaluation code. The paper describes $\Sigma q_t / T^*$ (visible frames only); the difference is $\approx$3--4\% on AntiUAV600. \\
\cmidrule{2-3}
& Template update disabled & Matches released checkpoint behaviour; intentional in the public codebase. \\
\midrule
\multirow{5}{*}{E2} & AugLy $\to$ librosa for audio noise & AugLy is incompatible with the Snellius CUDA environment (\texttt{libcudart.so} version mismatch). Librosa noise is demonstrably weaker than AugLy, contributing to the 9pp fusion accuracy gap over the paper. \\
\cmidrule{2-3}
& RF evaluated clean-only & Public release contains only pre-computed spectrograms; raw I/Q data unavailable for noise injection before STFT. Condition~(g) applies spectrogram-domain noise as a controlled alternative. \\
\cmidrule{2-3}
& VGG learning rate: $10^{-4}$ (paper: $10^{-3}$--$10^{-2}$) & VGG collapses to majority-class prediction at the paper's learning rate range. \\
\cmidrule{2-3}
& All checkpoints retrained from scratch & Published checkpoints (except LeNet) are corrupt. \\
\cmidrule{2-3}
& Four bug fixes applied$^*$ & (1) RF spectrogram augmentations removed (physically incorrect flips/rotations on pre-computed spectrograms); (2) ImageNet normalisation was computed but not applied in the video pipeline; (3) backbone weights were not frozen during fusion training; (4) non-deterministic training seeds. \\
\midrule
\multirow{6}{*}{E3} & Tracker results written only on GT frames & MM-UAV annotates at a $\sim$20-frame stride. The released frame guard had been removed in an earlier revision, so the tracker emitted detections on $\sim$2{,}000 frames/sequence against $\sim$110 annotated; TrackEval scores each as a false positive on empty GT ($\sim$438K FPs, MOTA $\approx -1500\%$). A GT-driven frame set (read from \texttt{gt.txt} at runtime) restores writing to annotated frames only. Prerequisite for any valid MOTA/HOTA. \\
\cmidrule{2-3}
& Det. threshold $0.5 \to 0.3$; full assoc. enabled & Matches the paper's \texttt{final\_def.sh}. Parser defaults (conf $0.5$, IoU-only matching) suppress recall and admit FPs on sparse-GT sequences, independently reproducing the $\approx -1500\%$ MOTA failure. \\
\cmidrule{2-3}
& CLI args propagated in multiprocessing & \texttt{process\_sequences()} regenerated default arguments per sequence, discarding all command-line tracker parameters; arguments are now deep-copied into each worker. Without this the tracker silently ran on defaults. \\
\cmidrule{2-3}
& Stage-1 weights loaded; import paths fixed & Published Stage-1 weights (\texttt{yolox\_s\_2\_stream}) are loaded and frozen, training only Stage-2 (detection heads, plus ADFM where retained). Packaging fixes (module rename, tar-aware dataset loader, tar key-prefix stripping) restore the repo's intended imports with no change to training logic. \\ \cmidrule{2-3} & Stage 2 retrained on a 1,080/120 split & MM-UAV has no validation split: its \texttt{val} annotation files cover the 121 test sequences, which the released trainer uses to select the Stage-2 checkpoint. All conditions, including the baselines, are retrained with selection on a 10\% tune split of the training sequences (seed 42). \\
\cmidrule{2-3}
& HPC configured for Snellius/Lustre & Stable execution required \texttt{num\_workers}{=}0, \texttt{pin\_memory}{=}False, CPU-first checkpoint loading, single-threaded OpenCV, a Lustre metadata warm-up, and extraction of the 78{,}738 annotation-referenced images from 1{,}321 tar shards. Workarounds for a CUDA/cv2 deadlock and Lustre I/O contention; correctness unaffected. \\
\bottomrule
\multicolumn{3}{l}{\footnotesize $^*$ The four fixes correct discrepancies between the paper's stated methodology and its published code.} \\
\end{tabular}
\end{table*}

\clearpage

\section{Extended Diagnostics}
\label{sec:apx:unc_diag}

\setcounter{table}{0}
\renewcommand{\thetable}{E.\arabic{table}}

\subsection{E2 Robustness and Calibration Checks}

\begin{table*}[htbp]
\centering
\caption{E2 temperature scaling ablation (top-label ECE, 15 bins). $T$ fitted by NLL minimisation on the seed-42 validation split; ECE is the mean over the three noise seeds. Models hitting $T = 20$ (search upper bound) exhibit extreme over-confidence. Only sigmoid baselines are temperature-scaled; EviDS conditions report raw evidential ECE.}
\label{tab:e2_tempscale}
\small
\begin{tabular}{l c c c c}
\toprule
\textbf{Model} & $T$ & \textbf{Baseline ECE} & \textbf{Temp-Scaled ECE} & $\Delta$\textbf{ECE} \\
\midrule
LeNet (Audio) & 8.11 & 0.323 & 0.090 & $-$0.233 \\
VGG (Audio) & 20.00$^*$ & 0.358 & 0.102 & $-$0.256 \\
ResNet-10 (Video) & 20.00$^*$ & 0.636 & 0.232 & \textbf{$-$0.404} \\
MobileNet (Video) & 20.00$^*$ & 0.321 & 0.148 & $-$0.173 \\
ResNet-10 (RF) & 3.52 & \textbf{0.149} & 0.066 & $-$0.083 \\
Late Fusion & 0.43 & 0.154 & 0.057 & $-$0.097 \\
GMU & 9.23 & 0.317 & \textbf{0.029} & $-$0.288 \\
\midrule
\multicolumn{5}{l}{\footnotesize EviDS fusion conditions (d)--(g) are not temperature-scaled; their raw ECE (three-seed mean, Table~\ref{tab:e2_results}) is 0.254 (d), 0.091 (e), 0.054 (f), 0.148 (d$'$) and 0.328 (g).} \\
\bottomrule
\multicolumn{5}{l}{\footnotesize $^*$ Hit optimisation upper bound; probabilities saturated near 0 or 1.} \\
\end{tabular}
\end{table*}

\begin{table*}[htbp]
\centering
\caption{Gating efficiency decomposition for condition (f) vs.\ (e) on the TRIDENT test set (seed 42). Condition (1) measures ideal throughput with full batching; (2) isolates the batching penalty by processing one sample at a time; (3) adds the gate on top of streaming but does \emph{not} skip encoder computation---the gate only zeros belief masses after the fact. The projected row subtracts per-modality encoder time weighted by observed skip rates and adds the measured gate-decision overhead.}
\label{tab:e2_efficiency}
\small
\begin{tabular}{l c c c c}
\toprule
\textbf{Condition} & \textbf{Mode} & \textbf{Latency (ms/sample)} & \textbf{Throughput (FPS)} & \textbf{Overhead} \\
\midrule
(1) Ungated DS & Batched & 0.24 & 4,200 & --- (ideal) \\
(2) Ungated DS & Streamed & 5.45 & 184 & +5.21 ms (22$\times$ batching penalty) \\
(3) Gated DS & Streamed (no encoder skip) & 5.71 & 175 & +0.26 ms (gate logic only) \\
\midrule
\multicolumn{5}{l}{\textbf{Projected with encoder skipping}} \\
\quad Gated DS & Streamed (true skip) & $\approx$4.17 & $\approx$240 & $-$1.28 ms ($-$23\% vs.\ ungated streamed) \\
\bottomrule
\multicolumn{5}{l}{\footnotesize Projected latency $=$ (2) $-$ $\Sigma$(per-modality time $\times$ skip rate) $+$ gate overhead.} \\
\multicolumn{5}{l}{\footnotesize Per-encoder times: Audio 2.10, Video 2.55, RF 0.56 ms; skip rates (seed 42): Audio 42.6\%, Video 21.4\%, RF 17.0\%; gate overhead $+$0.26 ms.} \\
\multicolumn{5}{l}{\footnotesize Encoder-time savings ($-$1.54 ms) are projected---the benchmark always runs all encoders; only belief masses are zeroed.} \\
\multicolumn{5}{l}{\footnotesize Realising the saving requires calling only the active encoders, which the benchmark does not implement.} \\
\end{tabular}
\end{table*}

\begin{table*}[htbp]
\centering
\caption{E2 clean-test reference results (1,320 TRIDENT test segments). Noisy rows restate the three-seed means $\pm$ sample standard deviation of Table~\ref{tab:e2_results}; clean rows disable test-time degradation for all modalities. Because the seed parameterises only the test-time noise transform, the clean pass is deterministic and identical across seeds; the zero std is by construction, not a measured coincidence. Conditions (d) and (e) converge to $\approx$99\% under clean conditions, confirming their noisy-test gaps are degradation-driven; the gated condition (f) reaches only 97.35\% because the gate fires even on clean inputs (see note).}
\label{tab:e2_clean_test}
\small
\begin{tabular}{l c c c}
\toprule
\textbf{Condition} & \textbf{Acc (\%)} $\uparrow$ & \textbf{Macro-F1} $\uparrow$ & \textbf{ECE} $\downarrow$ \\
\midrule
(d) EDL + Average (noisy)   & 97.20 $\pm$ 0.08 & 0.970 $\pm$ 0.001 & 0.254 $\pm$ 0.001 \\
(d) EDL + Average (clean)   & 98.79 $\pm$ 0.00 & 0.987 $\pm$ 0.000 & 0.127 $\pm$ 0.000 \\
\addlinespace
(e) EDL + DS (noisy)        & 97.53 $\pm$ 0.17 & 0.973 $\pm$ 0.002 & 0.091 $\pm$ 0.003 \\
(e) EDL + DS (clean)        & 99.02 $\pm$ 0.00 & 0.989 $\pm$ 0.000 & 0.016 $\pm$ 0.000 \\
\addlinespace
(f) EDL + DS + Gating (noisy) & 91.82 $\pm$ 1.55 & 0.911 $\pm$ 0.018 & 0.054 $\pm$ 0.007 \\
(f) EDL + DS + Gating (clean) & 97.35 $\pm$ 0.00 & 0.971 $\pm$ 0.000 & 0.033 $\pm$ 0.000 \\
\bottomrule
\end{tabular}
\vspace{4pt}
\begin{minipage}{\textwidth}
\footnotesize
Clean evaluation uses the same checkpoints as the noisy evaluation with test-time degradation disabled (Section~\ref{sec:e2}). Even on clean inputs, audio uncertainty exceeds $\tau_{\text{high}}$ on 11.7\% of segments, which keeps the audio stream gated off for 53.9\% of clean segments: under the canonical belief-mass construction these interventions cost accuracy (97.35\% vs.\ 99.02\%).
\end{minipage}
\end{table*}

\paragraph{Belief-mass construction (E2 robustness check).} Dempster--Shafer fusion maps each modality's evidential output to a basic probability assignment, and more than one valid assignment exists. The reported E2 results use the canonical belief-mass construction $b=e/S \equiv \hat{p}-u/K$ (Equation~\ref{eq:belief_mass}); an alternative valid construction, $b=\hat{p}(1-u)$, was also evaluated over the saved per-modality outputs. Both satisfy the basic-probability-assignment condition $\sum_k b_k^{(m)} + u^{(m)} = 1$, and because the per-modality heads are trained independently of the fusion rule, the comparison requires no retraining. The Dempster--Shafer-versus-average ordering reverses between the two constructions --- DS leads by $0.15$--$0.61$~pp under the canonical construction, the average by $0.56$~pp under the alternative --- but both margins fall within bootstrap noise: the canonical-construction margin is the same DS-vs-average comparison already tested in Section~\ref{sec:results} ($p=0.332$); the alternative-construction margin, newly tested here, gives 95\% CI $[-1.82, +3.03]$~pp ($p=0.32$). So the absence of a detectable difference between DS fusion and simple averaging (Section~\ref{sec:results}) does not depend on the choice of construction.

\subsection{Crop-Level Control (E3)}

\begin{table*}[htbp]
\centering
\caption{E3 crop-level control: vacuity UAUC over detection crops evaluated through the same frozen evidential model with no retraining. Crops use a $1.5\times$ context margin around each detection box, resized to $640\times640$ with aspect-preserving zero-padding, drawn from detections at the analysis floor (conf$=$0.001) on a 15\% random subsample of test-set image IDs (seed 42), a computational subsample within the 121 test sequences rather than a separate data split; UAUC over matched vs.\ unmatched labels is prevalence-invariant, so balancing is unnecessary. Aggregate CIs are sequence-clustered over the full 121 test sequences (resampled with repeats). Each row is one checkpoint; the same crop is fed to both streams and vacuity is read from the RGB head. Anchor-level reference values (all detections) from Table~\ref{tab:e3_results}.}
\label{tab:e3_crop_control}
\small
\begin{tabular}{l c c c c c}
\toprule
\textbf{Checkpoint} & \textbf{Anchor-level} & \textbf{Crop aggregate [95\% CI]} & \textbf{Small} & \textbf{Medium} & \textbf{Large} \\
\midrule
Averaging-trained (d) & 0.960 & 0.714 [0.684, 0.736] & 0.732 [0.700, 0.756] ($n{=}25{,}314$) & 0.619 ($n{=}1{,}089$) & 0.615 ($n{=}235$) \\
DS-trained (e) & 0.972 & 0.605 [0.554, 0.646] & 0.583 [0.527, 0.629] ($n{=}25{,}033$) & 0.688 ($n{=}1{,}164$) & 0.495 ($n{=}226$) \\
\bottomrule
\end{tabular}
\vspace{4pt}
\begin{minipage}{\textwidth}
\footnotesize
Scale strata use the experiment's bins on source-box area: small $<200$\,px$^2$, medium 200--1000\,px$^2$, large $>1000$\,px$^2$. The small-object stratum dominates both aggregates; the medium and large strata are reported descriptively ($n$ too small for clustered inference). The fine-tuning arm of the original control design is excluded: its training loss was identically zero (a $K{=}1$ loss-degeneracy bug, Section~\ref{sec:loss}), so no weights were updated.
\end{minipage}
\end{table*}

\subsection{Uncertainty-Ranking Diagnostics (E1, E3)}

\begin{table*}[htbp]
\centering
\caption{E1 detector-level uncertainty diagnostics (50 val sequences). Columns report UAUC of detector vacuity $u$ and of confidence $1-p_{\text{uav}}$ against false positives, and mean vacuity on correct detections versus false positives (115 and 98 TPs; 6 and 5 FPs). With only 5--6 false positives across all sequences, UAUC is not reliably estimable.}
\label{tab:e1_uncertainty_diag}
\begin{tabular}{l c c c c c}
\toprule
\textbf{Condition} & UAUC$_{u\to\text{FP}}$ & UAUC$_{(1-p)\to\text{FP}}$ & $\bar{u}$ (correct) & $\bar{u}$ (FP) & $N_{\text{FP}}$ \\
\midrule
EDL-270 & 0.54 & 0.76 & 0.44 & 0.45 & 6 \\
EDL-270 + $\tau_{\text{det}}$ & 0.58 & 0.82 & 0.41 & 0.43 & 5 \\
\bottomrule
\end{tabular}
\end{table*}

\begin{table*}[htbp]
\centering
\caption{E1 detector vacuity by challenge attribute (EDL-270). Means are computed over all detector-active frames of the 18 attribute-annotated validation sequences ($\sim$12K frames, predominantly background), not over matched detections; on frames without a detection the recorded value is the mean vacuity over all anchors of the frame. These means therefore describe whole-frame, mostly background evidence rather than detection uncertainty; all effect sizes are negligible ($|d|<0.15$) and no difficulty signal can be inferred.}
\label{tab:e1_attr_detector}
\begin{tabular}{l c c c c c}
\toprule
\textbf{Attribute} & $\bar{u}$ (inactive) & $\bar{u}$ (active) & $\Delta u$ & Cohen's $d$ & $N_{\text{active}}$ \\
\midrule
Occlusion & 0.0210 & 0.0186 & $-$0.0024 & $-$0.072 & 138 \\
Fast Motion & 0.0206 & 0.0231 & $+$0.0025 & $+$0.075 & 1\,831 \\
Scale Variation & 0.0210 & 0.0193 & $-$0.0017 & $-$0.050 & 43 \\
IR Crossover & 0.0238 & 0.0203 & $-$0.0035 & $-$0.085 & 2\,077 \\
Dynamic Background Clusters & 0.0207 & 0.0223 & $+$0.0016 & $+$0.048 & 2\,178 \\
Target Scale & 0.0200 & 0.0245 & $+$0.0045 & $+$0.110 & 2\,332 \\
\bottomrule
\end{tabular}
\end{table*}

\begin{table*}[htbp]
\centering
\caption{E3 RGB detection--tracking decomposition across confidence thresholds (MM-UAV test set, 121 sequences, $\text{GT}=28{,}017$ boxes). Condition~(c) is reported at three thresholds to examine the effect of the confidence threshold: correct detections are present at the analysis floor ($\text{conf}=0.001$), but almost none survive tracking at $\text{conf}=0.1$ (c$'$, which lowers only the detection threshold; new tracks still require a score above 0.7) or $0.3$. The baseline and the decision-level fusion condition~(d) are shown at the operating threshold for reference. The (d) versus (a) comparison shows that averaging improves recall (fewer false negatives) but adds false positives and identity switches, and the false-positive cost drives the MOTA gap. MOTA is not reported at the analysis floor (conf${=}0.001$ is a detection-count diagnostic, not an operating point; $\sim$16{,}500 false positives would dominate any meaningful tracking metric).}
\label{tab:e3_confsweep}
\small
\begin{tabular}{l c c c c c c}
\toprule
\textbf{Condition} & \textbf{conf} & \textbf{TP} $\uparrow$ & \textbf{FP} $\downarrow$ & \textbf{FN} $\downarrow$ & \textbf{Recall} $\uparrow$ & \textbf{MOTA} $\uparrow$ \\
\midrule
(c) Evidential + ADFM        & 0.001 & 23{,}743 & 16{,}483 & --- & --- & ---\textsuperscript{$\dagger$} \\
(c$'$) Evidential + ADFM     & 0.1   & 80       & 42      & 27{,}937 & 0.3\%  & 0.1\% \\
(c) Evidential + ADFM        & 0.3   & 30       & 23      & 27{,}987 & 0.1\%  & 0.0\%  \\
\midrule
(a) MMA-SORT (baseline)      & 0.3   & 21{,}993 & 3{,}839 & 6{,}024  & 78.5\% & 64.3\% \\
(d) Evidential + Average     & 0.3   & 22{,}646 & 6{,}690 & 5{,}371  & 80.8\% & 55.9\% \\
\bottomrule
\multicolumn{7}{l}{\footnotesize \textsuperscript{$\dagger$}Analysis-floor row counts detections with $\text{IoU}\geq0.5$ to any GT box (no tracker; not one-to-one, so FN and recall are undefined); all other rows are TrackEval outputs.} \\
\end{tabular}
\end{table*}

\vspace{3em}

\subsection{Reliability and Gate-Sweep Figures (E1, E2)}

\setcounter{figure}{0}
\renewcommand{\thefigure}{E.\arabic{figure}}

\begin{figure*}[htbp]
\centering
\includegraphics[width=\textwidth]{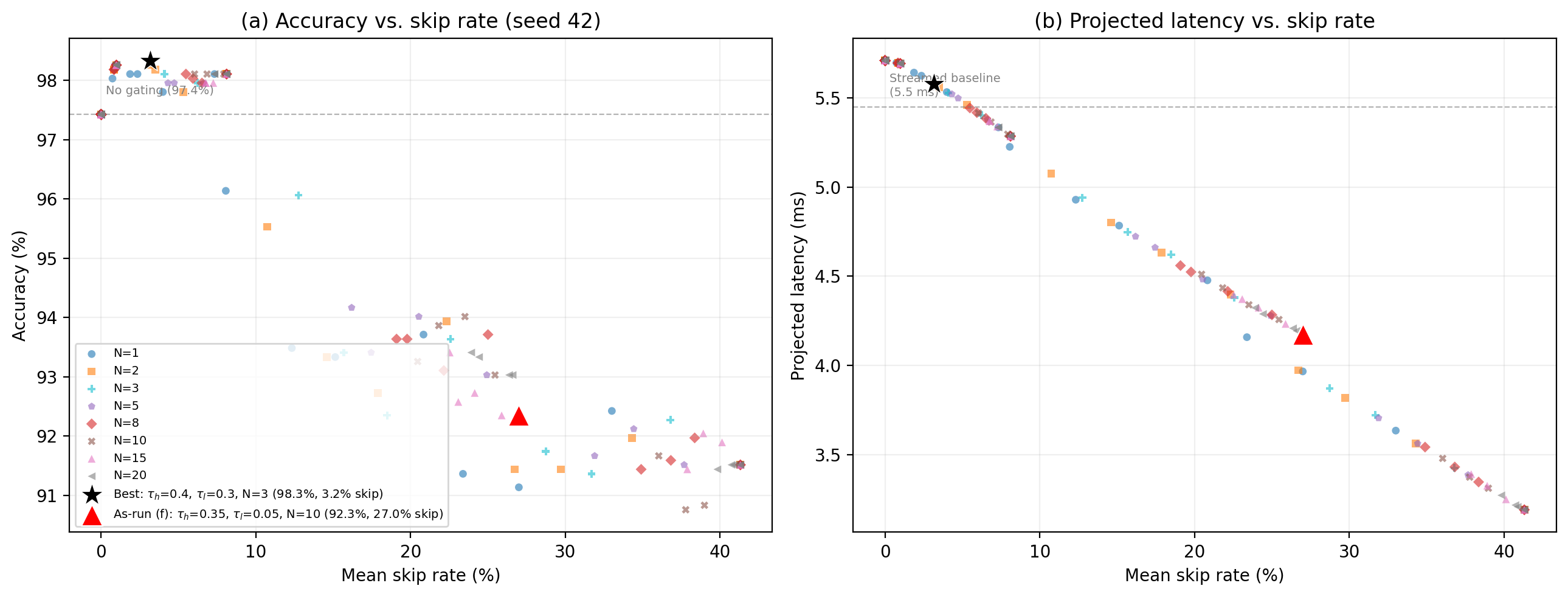}
\caption{Temporal gate hyperparameter sweep (seed 42, 1,320 test segments; 664 configurations), evaluated offline over cached encoder outputs under the canonical belief-mass construction. Marker shapes denote the suppression window $N$; the grey dashed line is the ungated DS baseline. (a)~Accuracy versus mean skip rate, which is bimodal --- only low-skip configurations preserve accuracy ($\star$ = best configuration; $\blacktriangle$ = as-run condition~(f)). (b)~Projected end-to-end latency versus skip rate; savings are projected, as the benchmark evaluates every encoder before the gate decision (Table~\ref{tab:e2_efficiency}). The best configuration ($\star$) was selected on the test split.}
\label{fig:e2_gate_sweep}
\end{figure*}

\begin{figure*}[htbp]
\centering
\includegraphics[width=\textwidth]{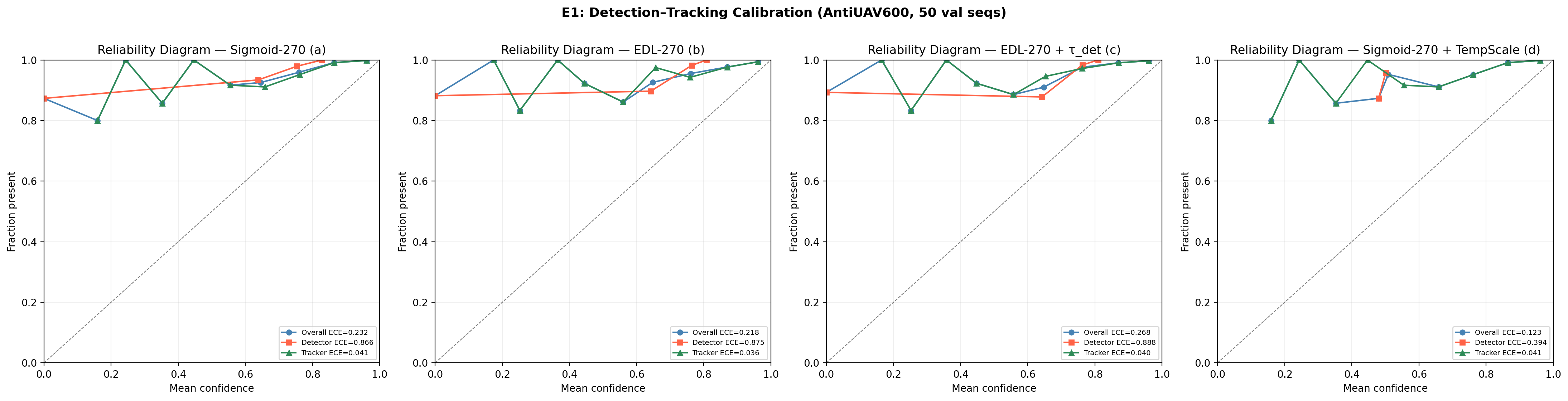}\\[6pt]
\includegraphics[width=\textwidth]{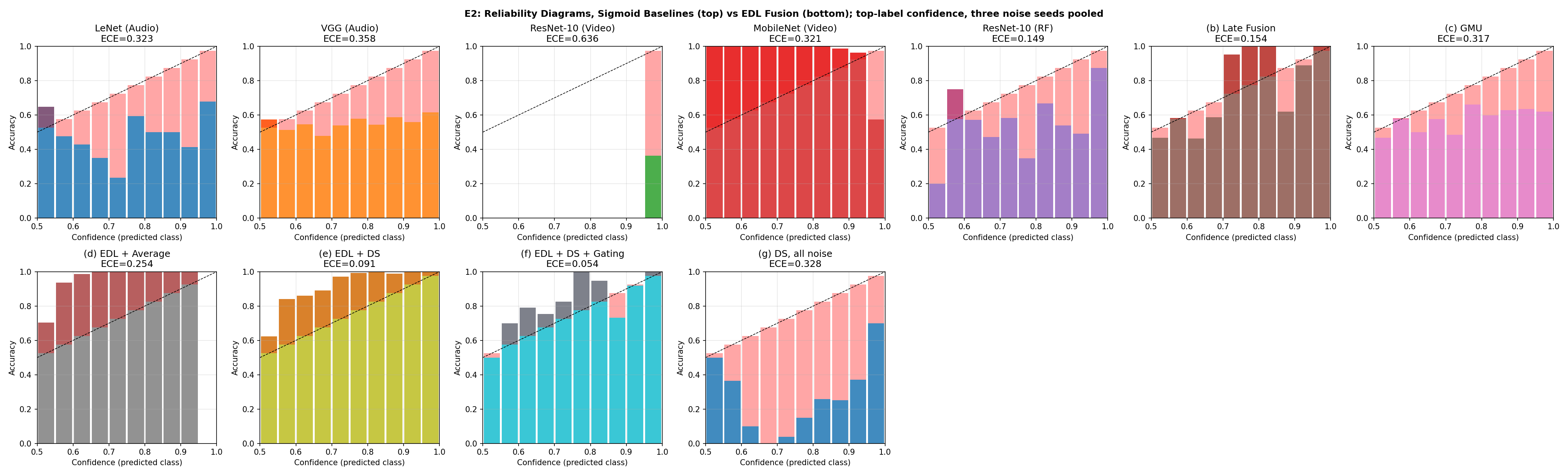}
\caption{Reliability diagrams. \textbf{Top, E1} (left to right): conditions (a)~Sigmoid-270, (b)~EDL-270, (c)~EDL-270$+\tau_{\text{det}}$, and (d)~Sigmoid-270$+$TempScale ($T\!=\!200$), each showing overall, detector, and tracker curves; the detector curves are dominated by detector-active frames without a detection (confidence 0), so they mainly reflect the miss rate, and in (d) temperature scaling at its search bound maps all detector confidences to $\approx 0.5$. \textbf{Bottom, E2} (top-label confidence; segments of the three noise seeds pooled, ECE is the three-seed mean; noisy test protocol): upper row, the five unimodal backbones and the baseline fusion heads (Late Fusion, GMU) --- the ResNet-10 (Video) panel concentrates all predictions in the top-confidence bin at 36\% accuracy, a constant-prediction collapse to the minority class (Appendix Table~\ref{tab:e2_results_full}); lower row, the evidential fusion conditions (d), (e), (f) and (g): averaging (d) is under-confident, DS (e) and gated DS (f) lie close to the diagonal, and the all-noise foil (g) is over-confident. Bars show accuracy per confidence bin; the shaded part is the gap to the diagonal. Curves or bars above the diagonal indicate under-confidence; below, over-confidence.}
\label{fig:apx:reliability}
\end{figure*}

\clearpage
\setcounter{figure}{0}
\renewcommand{\thefigure}{F.\arabic{figure}}

\begin{landscape}
\section{Methodology Overview}\label{appendix:methodology}

\vspace*{\fill}
\begin{figure}[H]
    \centering
    \includegraphics[width=0.96\linewidth,height=0.78\textheight,keepaspectratio]{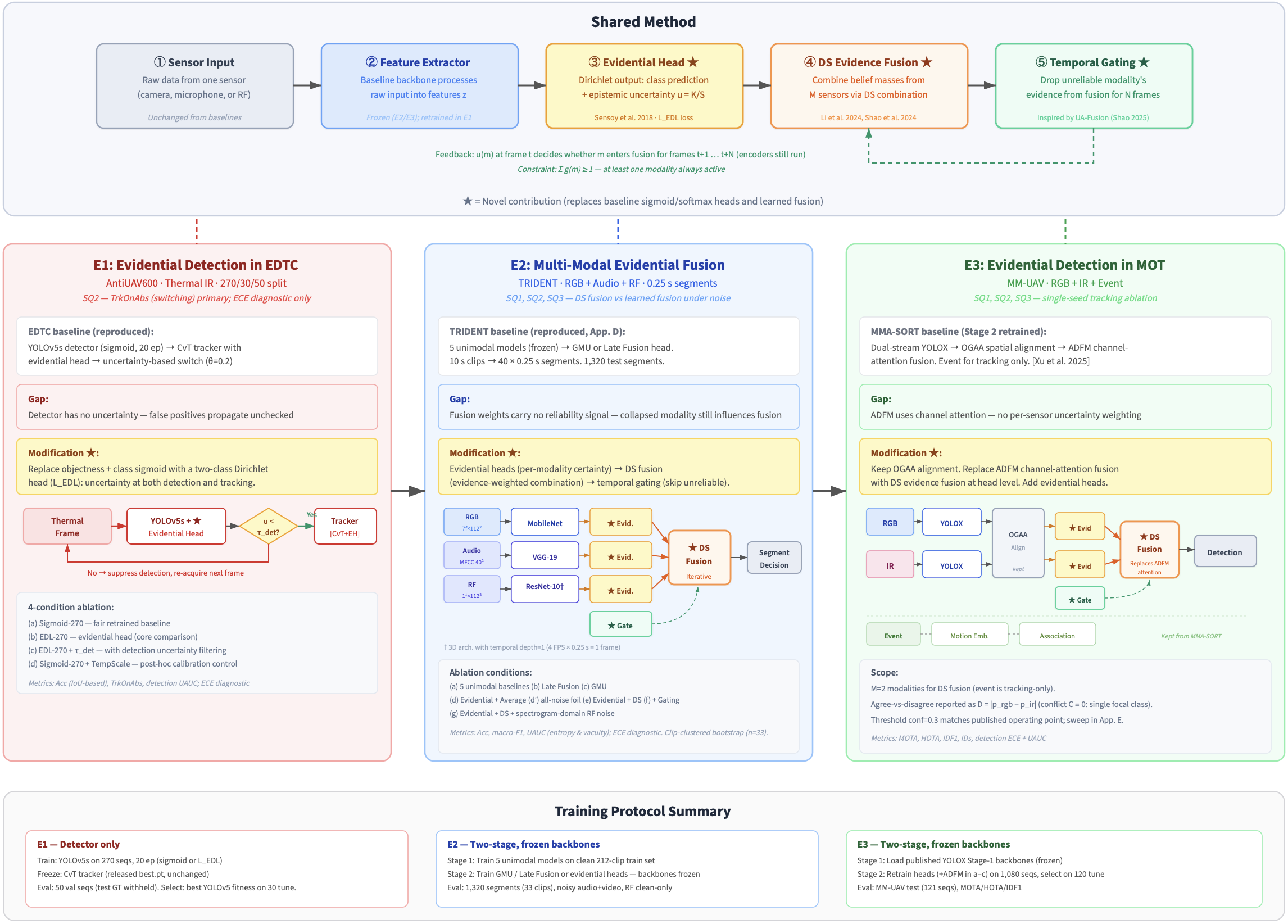}
    \caption{Framework overview of the general evidential pipeline and three experiments testing individual components.}
    \label{fig:framework}
\end{figure}
\vspace*{\fill}
\end{landscape}

\FloatBarrier
\clearpage
\end{document}